%% file: main.tex
\documentclass[lettersize,journal]{IEEEtran}
\usepackage{graphics}
\usepackage{amsmath,amsfonts}
\usepackage{algorithmic}
\usepackage{algorithm}
\usepackage[caption=false,font=normalsize,labelfont=sf,textfont=sf]{subfig}
\usepackage{textcomp}
\usepackage{stfloats}
\usepackage{url}
\usepackage{verbatim}
\usepackage{graphicx}
\usepackage{cite}
\usepackage{url}
\usepackage{times}
\usepackage{epsfig}
\usepackage{amssymb}
\usepackage{colortbl}
\usepackage{color}
\usepackage{threeparttable}
\usepackage{booktabs}
\usepackage{array}
\usepackage{adjustbox}
\usepackage[normalem]{ulem}
\usepackage{tabularx}
\usepackage[misc]{ifsym}
\usepackage[table,x11names]{xcolor}

\usepackage{colortbl}
\usepackage{enumitem}
\usepackage{multirow}
\usepackage{rotating}
\usepackage{dsfont}
\usepackage{makecell}
\usepackage{bm}
\usepackage{xspace} 

\usepackage{ragged2e} 
\usepackage{siunitx}
\usepackage[breaklinks=true,colorlinks,citecolor=blue,urlcolor=blue,linkcolor=blue,bookmarks=false]{hyperref}

\newcommand{\onedot}{.\xspace}

\usepackage{pifont}
\newcommand{\cmark}{\ding{51}}%
\newcommand{\xmark}{\ding{55}}%

\newcolumntype{C}[1]{>{\centering\arraybackslash}p{#1}}
\definecolor{customred}{HTML}{B32B0D}
\definecolor{customgreen}{RGB}{0,150,0}
\usepackage{orcidlink}
\begin{document}

\title{Category Discovery: An Open-World Perspective}

\author{ Zhenqi He,
        Yuanpei Liu,
        Kai Han 
\thanks{Z.~He, Y.~Liu, and K.~Han are with the School of Computing and Data Science, The University of Hong Kong, Hong Kong SAR, China. Corresponding author: Kai Han (kaihanx@hku.hk)}
}



\maketitle
\def\@onedot{\ifx\@let@token.\else.\null\fi\xspace}
\def\eg{\emph{e.g}\onedot} \def\Eg{\emph{E.g}\onedot}
\def\ie{\emph{i.e}\onedot} \def\Ie{\emph{I.e}\onedot}
\def\cf{\emph{c.f}\onedot} \def\Cf{\emph{C.f}\onedot}
\def\etc{\emph{etc}\onedot} \def\vs{\emph{vs}\onedot}
\def\wrt{w.r.t\onedot} \def\dof{d.o.f\onedot}
\def\etal{\emph{et al}\onedot}
\input{Secs/0_abstract}

\IEEEdisplaynontitleabstractindextext
\IEEEpeerreviewmaketitle

\input{Secs/1_intro}
\input{Secs/2.1_background}
\input{Secs/3_categorization}

\input{Secs/4_benchmark}
\input{Secs/5_Future}

\input{Secs/6_conclusion}

\begin{tiny}
  {
  \bibliographystyle{IEEEtran}
  \normalem
  \bibliography{main}
  }
\end{tiny}

\clearpage
\begin{appendices}
\input{Secs/appendix}

\end{appendices}

\vfill

\end{document}

%% file: Secs/0_abstract.tex
\begin{abstract}
Category discovery (CD) is an emerging open-world learning task, which aims at automatically categorizing unlabelled data containing instances from unseen classes, given some labelled data from seen classes.
This task has attracted significant attention over the years and leads to a rich body of literature trying to address the problem from different perspectives. 
In this survey, we provide a comprehensive review of the literature, and offer detailed analysis and in-depth discussion on different methods. 
Firstly, we introduce a taxonomy for the literature by considering two base settings, namely novel category discovery (NCD) and generalized category discovery (GCD), and several derived settings that are designed to address the extra challenges in different real-world application scenarios, including continual category discovery, skewed data distribution, federated category discovery, etc. 
Secondly, for each setting, we offer a detailed analysis of the methods encompassing three fundamental components, representation learning, label assignment, and estimation of class number. 
Thirdly, we benchmark all the methods and distill key insights showing that large-scale pretrained backbones, hierarchical and auxiliary cues, and curriculum-style training are all beneficial for category discovery, while challenges remain in the design of label assignment, the estimation of class numbers, and scaling to complex multi-object scenarios.
Finally, we discuss the key insights from the literature so far and point out promising future research directions. 
We compile a living survey of the category discovery literature at \href{https://github.com/Visual-AI/Category-Discovery}{https://github.com/Visual-AI/Category-Discovery}.
\end{abstract}


%% file: Secs/1_intro.tex
\vspace{-10pt}

\section{Introduction}
\label{sec:intro}

\begin{figure}[t!]
\centering
\setlength{\abovecaptionskip}{0pt} 
\includegraphics[width=0.48\textwidth]{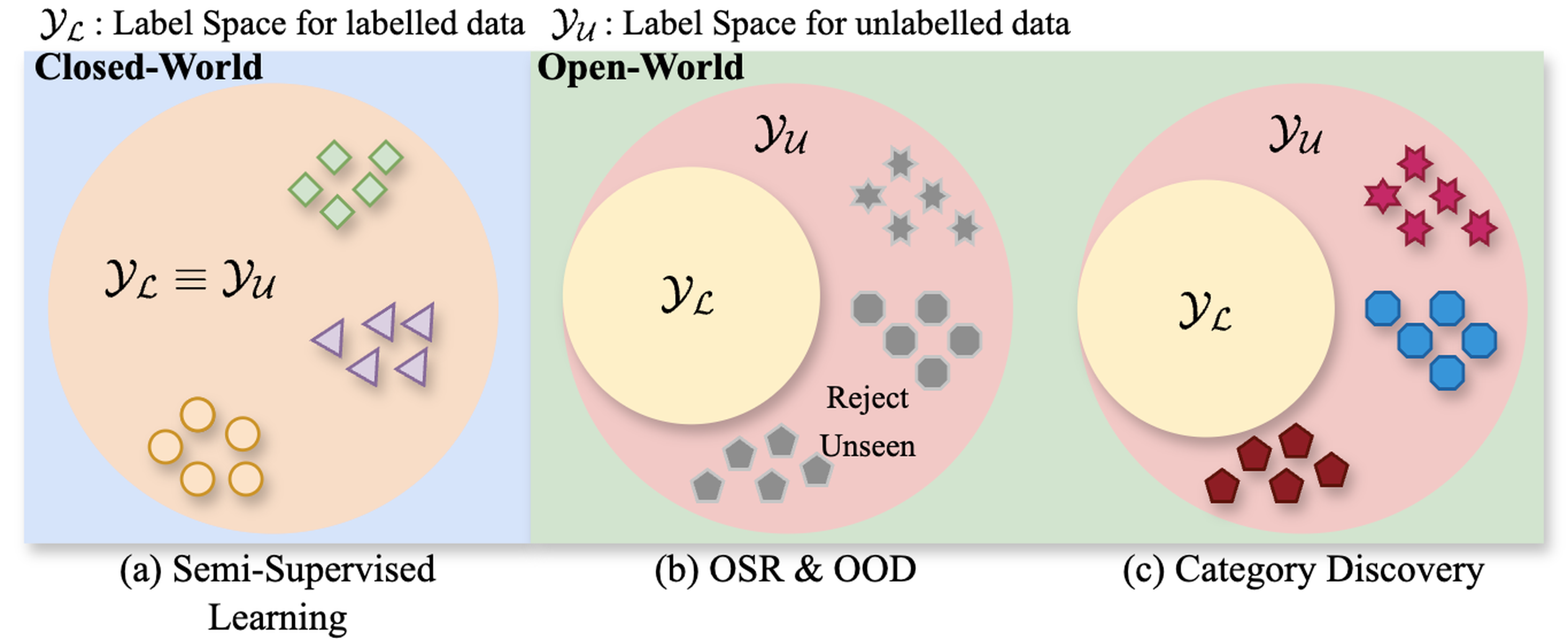}
\caption{ 
Comparison of category discovery with semi-supervised learning and OSR\&OOD.
}
\label{fig:intro}
\vspace{-16pt}
\end{figure}

\IEEEPARstart{D}{eep} learning algorithms have demonstrated exceptional performance, often surpassing human performance in tasks such as image classification~\cite{resnet,liu2021swin}, object detection~\cite{tan2020efficientdet,Meta-DETR-2022}, and semantic segmentation~\cite{ronneberger2015u,segementationSurvey2022}. However, these methods typically operate under a closed-world assumption, relying heavily on large amounts of labelled data and predefined labels, which limits their applicability in real-world scenarios. 
Semi-supervised learning~\cite{sslSurvey2022, dssl2018nips} alleviates this limitation by utilizing a small amount of labelled data alongside a large pool of unlabelled data, reducing the reliance on costly annotations. 
Yet, semi-supervised learning remains restricted by the assumption of fixed categories. 
In contrast, Open-World Learning allows models to encounter, recognize, and incorporate unseen classes beyond the predefined label set.
{
Among open-world learning tasks, Open-Set Recognition (OSR)~\cite{geng2020recent,osr2013} and Out-Of-Distribution (OOD) Detection~\cite{yang2022openood,yang2021oodsurvey} focus on flagging novel samples as ``unknown'' but stop short of grouping them, while Category Discovery goes further by clustering the entire unlabelled dataset into semantically coherent groups.}
%
%
As illustrated in Fig.~\ref{fig:intro}, Category Discovery (CD) distinguishes itself from semi-supervised learning and OSR\&OOD by clustering unlabelled data that includes unseen categories. 
{
Drawing inspiration from how humans classify new species based on existing knowledge, CD has proven valuable in many real-world scenarios.
For instance, in botanical research, taxonomists use labelled specimens of known species to organize newly collected unlabelled samples into existing taxa or to reveal novel species~\cite{karbstein2024species, kutuzova2024taxometer}, mirroring the core idea of CD: leverage knowledge from annotated data to cluster and discover both known and previously unknown categories.}
%
%

In recent years, CD has garnered increasing attention, leading to a proliferation of research exploring various methodologies and settings. 
CD is initially introduced as Novel Category Discovery (NCD) in \cite{Han2019DTC} to cluster unlabelled novel categories by leveraging knowledge from labelled base categories. 
{It is then extended to Generalized Category Discovery (GCD)~\cite{vaze2022gcd}, which removes the assumption that unlabelled samples only come from novel classes.  Instead, GCD accommodates a mixture of both seen and unseen categories in unlabelled data, aligning the task more closely with real-world scenarios.}
{Further advancing category discovery, recent methods have explored a wide range of challenging and realistic settings.
%
%
Continual Category Discovery (CCD)~\cite{zhang2022growmergeunifiedframework, zhao2023incremental} addresses incremental learning over time, while Federated Category Discovery (FCD)~\cite{pu2023federatedgeneralizedcategorydiscovery} focuses on decentralized model training under privacy constraints. Semantic Category Discovery (SCD)~\cite{han2024whatsnameclassindices} targets assigning semantic labels from an open vocabulary to unlabelled data. Other efforts tackle few-shot scenarios~\cite{chi2022meta}, long-tailed distributions~\cite{li2024generalizedcategoriesdiscoverylongtailed}, and domain shifts~\cite{wang2024hilolearningframeworkgeneralized}, enhancing the practicality and robustness of CD methods in real-world applications.}

\begin{figure*}[t]
\setlength{\abovecaptionskip}{-6pt} 
\centering
\includegraphics[width=1.0\textwidth]{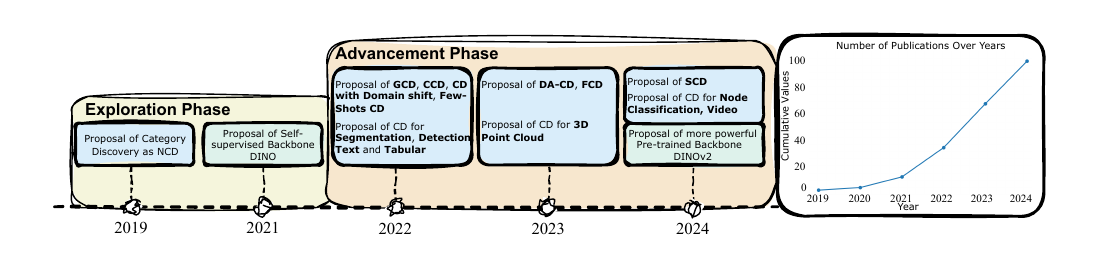}
\caption{ 
Roadmap and the cumulative number of publications for the development of Category Discovery from 2019 to 2024.
}
\vspace{-12pt}
\label{fig:number}
\end{figure*}

With the rapid advancements in CD, it is increasingly important to track and evaluate the latest advancements in this area.
As illustrated in Fig.~\ref{fig:number}, CD research has surged since 2022 across diverse scenarios. 
%
%
This surge motivates us to provide a thorough overview of existing methods and settings within the CD community, along with comprehensive evaluation results on widely used datasets through standardized reimplementations, facilitating fair comparisons.
%
{Although previous surveys~\cite{zhu2024open, troisemaine2023novelclassdiscoveryintroduction} have discussed CD, \cite{troisemaine2023novelclassdiscoveryintroduction} focuses predominantly on NCD, with limited attention to broader and more practical scenarios including GCD, continual category discovery, and other related settings, and \cite{zhu2024open} presents a review of open-world machine learning, framing category discovery as a central component alongside unknown rejection and continual learning.}
%
We provide a focused and in-depth analysis specifically on \textit{Category Discovery}, including systematic analysis of category discovery methods across multiple scenarios, and providing a holistic overview of the current landscape and highlighting future research directions.

\noindent
\textit{\textbf{Contributions.}} 
{In this survey, we first provide a systematic review of the current Category Discovery literature with a comprehensive taxonomy.
Particularly, we group methods into  \textit{Novel Category Discovery} and \textit{Generalized Category Discovery} based on their assumptions on the unlabelled data---we denote this as {\textit{base settings}}; 
Building on the base settings, we introduce seven derived settings that adapt category discovery to more practical conditions, including continual learning, decentralized federated training, domain shifts, imbalanced data distributions.}
%
To highlight the broader applicability of Category Discovery, we also examine recent advancements extending category discovery beyond visual data. 
Finally, we conclude with benchmark comparisons across standard datasets, {highlighting comparative strengths and limitations to provide actionable insights for future research}.
%

\begin{figure}[t]
	\centering

\setlength{\abovecaptionskip}{-4pt}
\includegraphics[width=1\linewidth]{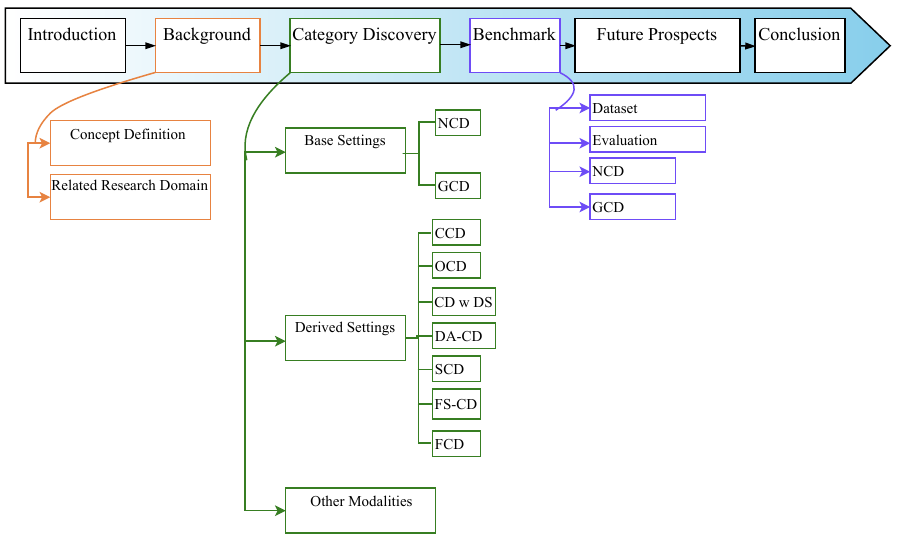}
    \put(-235,137){\tiny{\S\ref{sec:intro}}} 
    \put(-198,137){\tiny{\S\ref{Sec:background}}}
    \put(-153,137){\tiny{\S\ref{Sec:Methods}}}
    \put(-111.8,137){\tiny{\S\ref{Sec:benchmark}}}
    \put(-72,137){\tiny{\S\ref{Sec:future}}}
    \put(-32.5,137){\tiny{\S\ref{Sec:conclusion}}}
    \put(-221,110){\tiny{\S\ref{Subsec:concept_def}}}
    \put(-221,95){\tiny{\S\ref{Subsec:relatedDomains}}}
    \put(-160,107.7){\tiny{\S\ref{Subsec:concept}}}
    \put(-160,55){\tiny{\S\ref{Subsec:Scenario}}}
    \put(-158,6.5){\tiny{\S\ref{Subsec:beyondImages}}}
    \put(-71,122){\tiny{\S\ref{Subsec:dataset}}}
    \put(-68,113.5){\tiny{\S\ref{SubSec:eva_proto}}}
    \put(-75.8,105){\tiny{\S\ref{Subsec:NCD_benchmark}}}
    \put(-75.8,94.7){\tiny{\S\ref{Subsec:GCD_benchmark}}}
	\caption{Overview of this survey.}
	\label{fig:overview}
    \vspace{-18pt}
\end{figure}
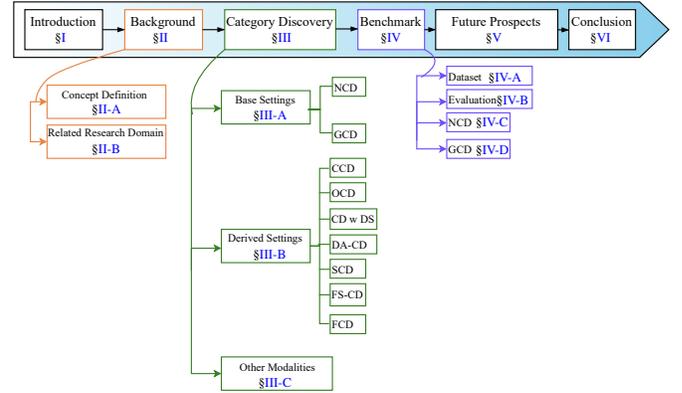

\noindent
\textit{\textbf{Outline.}} Fig.~\ref{fig:overview} outlines the structure of this survey. 
First, we provide a detailed analysis of problem definitions and related research domains in Sec.~\ref{Sec:background}. 
After that, we present a detailed taxonomy for both base and derived settings in Sec.~\ref{Sec:Methods}. 
We then move to Sec.~\ref{Sec:benchmark}, which introduces commonly used datasets in CD and evaluation protocols, followed by a performance comparison of these methods. 
Finally, we explore open questions and potential future research directions in Sec.~\ref{Sec:future}, culminating with a conclusion in Sec.~\ref{Sec:conclusion}.

%% file: Secs/2.1_background.tex
\vspace{-5pt}

\section{Background}
\label{Sec:background}
\noindent
\textit{Overview:}  {This section begins with a clarification of the definitions for base Category Discovery settings }in Sec.~\ref{Subsec:concept_def}, followed by a comparison with related research areas including clustering, Semi-Supervised Learning (SSL), Zero-Shot Learning (ZSL), Out-Of-Distribution  (OOD) Detection, Open-Set Recognition (OSR), and Open Vocabulary Learning in Sec.~\ref{Subsec:relatedDomains} with a summary provided in Tab.~\ref{tab:conceptComparison}.

\subsection{Problem Statements}
\label{Subsec:concept_def}
In the subsequent context, we denote the entire dataset as $\mathcal{D}$, which consists of a labelled subset $\mathcal{D}_L$ and an unlabelled subset $\mathcal{D}_U$. The corresponding data samples are represented by $\mathcal{X}$, with $\mathcal{X}_L$ and $\mathcal{X}_U$ denoting the labelled and unlabelled portions, respectively. The label spaces associated with the labelled and unlabelled sets are denoted by $\mathcal{Y}_L$ and $\mathcal{Y}_U$. We refer to the sets of base and novel categories as $\mathcal{C}_\mathcal{B}$ and $\mathcal{C}_\mathcal{N}$.

\begin{figure*}[ht]
\centering
\setlength{\abovecaptionskip}{-4pt}
\includegraphics[width=1\textwidth]{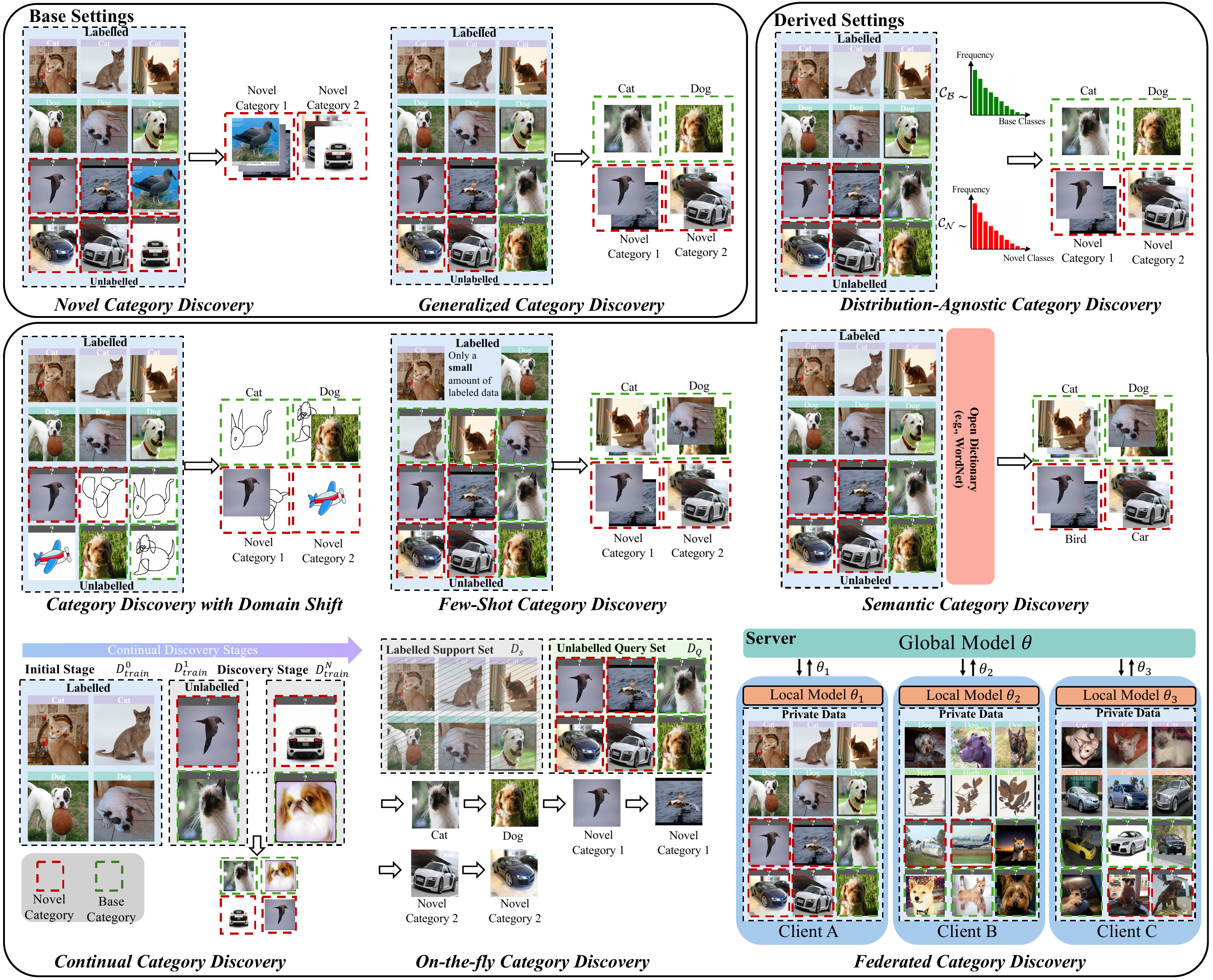}
\caption{ 
Overview of the category discovery taxonomy for both {\textit{base settings} and \textit{derived settings}}. {The base settings contain Novel Category Discovery, where only novel categories are present in the unlabelled set, and Generalized Category Discovery, where the unlabelled set contains both base and novel categories. For \textit{derived settings}:} Distribution-Agnostic Category Discovery: Existence of imbalanced data distribution; Category Discovery with Domain Shift: Domain shifts within data;  Few-Shot Category Discovery: Only a small number of labelled data is available; Semantic Category Discovery: Assignment of semantic labels of unlabelled data; Continual Category Discovery: Discovering novel categories continually; {On-the-fly Category Discovery:  Instantaneously identifying novel categories under inductive learning;} Federated Category Discovery: decentralized training across clients. 
}
\label{fig:settings}
\vspace{-15pt}
\end{figure*}

\hfill \break
\textbf{Novel Category Discovery (NCD).} NCD aims to transfer the knowledge learned from base categories to cluster unlabelled unseen categories. 
This draws inspiration from the observation that \textit{a child can readily differentiate between novel categories, such as birds and elephants, after mastering the classification of base categories like dogs and cats}~\cite{Han2019DTC}. 
Formally, given a dataset $\mathcal{D} = \mathcal{D}_L \cup \mathcal{D}_U$, where the labelled portion is $\mathcal{D}_L = \{(\mathbf{x}_i, y_i)\}_{i=1}^M \subset \mathcal{X} \times \mathcal{Y}_L$ and the unlabelled portion is $\mathcal{D}_U = \{(\mathbf{x}_i, \hat{y}_i)\}_{i=1}^K \subset \mathcal{X} \times \mathcal{Y}_U$ (with the labels $\hat{y}_i$ being inaccessible during training), the objective of NCD is to leverage the discriminative information learned from the annotated data to cluster the unlabelled data. It presumes that the label spaces of the labelled and unlabelled data are disjoint, \ie, $\mathcal{Y}_L \cap \mathcal{Y}_U = \varnothing$, while also assuming a high degree of semantic similarity between the base and novel categories.

\noindent 
\textbf{Generalized Category Discovery (GCD).} Extending the NCD paradigm, Generalized Category Discovery~\cite{vaze2022gcd} relaxes the assumption of disjoint label spaces between base and novel categories, offering a more realistic and challenging scenario.
In GCD, the labelled and unlabelled datasets share common categories, \ie, $\mathcal{Y}_L \cap \mathcal{Y}_U \neq \varnothing$, where the unlabelled data contains both seen and unseen categories.
This broader formulation aligns closely with real-world applications, such as plant species discovery, where an existing database of known species is expanded with newly discovered species, requiring the clustering of both seen and unseen instances.

Notably, an equivalent formulation has been introduced by Cao~\etal~\cite{orca2022} under the designation of \textit{Open-World Semi-Supervised Learning}. We uniformly use the term \textit{Generalized Category Discovery} in this survey.


\subsection{Related Research Problems}
\label{Subsec:relatedDomains}
\hfill \break
\textbf{Transfer Learning.} 
Transfer learning leverages knowledge from an annotated source dataset to improve performance on a target dataset by fine-tuning a pre-trained model~\cite{transferLearning2020,tlSurvey2010}. This paradigm enables the adaptation of learned representations to new tasks or domains, enhancing generalization when annotated data is available.

\noindent
\textbf{Semi-Supervised Learning (SSL).} SSL reduces the need for extensive annotations by training on a small labelled set and a large unlabelled set~\cite{sslSurvey2022,dssl2018nips}. In contrast to CD, SSL assumes a closed-world setting where both sets share the same categories, \ie, $\mathcal{Y}_L \equiv \mathcal{Y}_U$.

\noindent
\textbf{Out-Of-Distribution (OOD) Detection  \& Open-Set Recognition (OSR).} OOD and OSR address open-world challenges by detecting and rejecting test samples that differ from the training distribution~\cite{yang2021oodsurvey,osr2013,yang2022openood,geng2020recent}. 
{Particularly, OSR focuses on classifying instances from known categories while rejecting instances from  unseen classes, whereas OOD detection focuses on a more general problem of identifying instances deviating from the training data distribution.
}

\noindent
\textbf{Open Vocabulary Learning (OVL).} OVL addresses the classification of both base and novel categories by integrating auxiliary vision-aware language vocabularies. 
This paradigm is typically trained using data-label-vocabulary triplets, for example, $\mathcal{D}_{train} = \{ (\mathcal{X}_L, \mathcal{Y}_L, \mathcal{V}) \}$, where $\mathcal{V}$ represents auxiliary vocabulary information drawn from an extensive vocabulary space~\cite{wu2023open, zhu2024survey}. 
{The primary focus of this task is to leverage pretrained models with aligned visual and textual representations, thereby improving generalization in open-world scenarios.}

\noindent
\textbf{Clustering.} Clustering partitions data into groups based on inherent similarities, without relying on predefined labels or classification criteria~\cite{clusteringSurvey}. It typically employs distance or similarity measures to reveal latent structures, facilitating the discovery of natural groupings without external annotations.

\noindent
\textbf{Zero-Shot Learning (ZSL).} ZSL  enables recognition of novel categories unseen during training by leveraging auxiliary semantic information~\cite{zslSurvey2019,zsl2009}. It links representations of base categories to novel ones using semantic attributes, textual descriptions, or other embeddings, which serve as the bridge between seen and unseen classes.

\input{tabs/concept_comparison}

%% file: tabs/concept_comparison.tex
\begin{table}[]
\caption{Comparison of NCD and GCD with related tasks.}
\centering
\label{tab:conceptComparison}
\scalebox{1}{\begin{tabular}{l||ccc|cc}
\hline
\rowcolor[HTML]{D3D3D3}\multirow{2}{*}{} & \multicolumn{2}{c|}{Training}                                                  & \multicolumn{2}{c}{Testing}      \\ \rowcolor[HTML]{D3D3D3}
                        & \multicolumn{1}{c}{Labelled} & \multicolumn{1}{c|}{Unlabelled}  & \multicolumn{1}{c}{Novel} & Base \\ \hline
SSL         & \multicolumn{1}{c}{\textcolor{customgreen}{\ding{51}}}   & \multicolumn{1}{c|}{\textcolor{customgreen}{\ding{51}}}      & \multicolumn{1}{c}{\textcolor{customred}{\ding{55}}}      &   \textcolor{customgreen}{\ding{51}}   \\ 
OOD \& OSR  & \multicolumn{1}{c}{\textcolor{customgreen}{\ding{51}}}   & \multicolumn{1}{c|}{\textcolor{customred}{\ding{55}}}       & \multicolumn{1}{c}{Reject}      &   \textcolor{customgreen}{\ding{51}}   \\ 
Clustering  & \multicolumn{1}{c}{\textcolor{customred}{\ding{55}}}   & \multicolumn{1}{c|}{\textcolor{customred}{\ding{55}}}    & \multicolumn{1}{c}{\textcolor{customgreen}{\ding{51}}}      &   \textcolor{customred}{\ding{55}}   \\ 
ZSL  & \multicolumn{1}{c}{\textcolor{customgreen}{\ding{51}}}   & \multicolumn{1}{c|}{\textcolor{customred}{\ding{55}}}    & \multicolumn{1}{c}{\textcolor{customgreen}{\ding{51}}}      &   \textcolor{customred}{\ding{55}}   \\ 
OVL  & \multicolumn{1}{c}{\textcolor{customgreen}{\ding{51}}}   & \multicolumn{1}{c|}{\textcolor{customred}{\ding{55}}}    & \multicolumn{1}{c}{\textcolor{customgreen}{\ding{51}}}      &   \textcolor{customgreen}{\ding{51}}   \\ 
\hline
\rowcolor{yellow!15} NCD         & \multicolumn{1}{c}{\textcolor{customgreen}{\ding{51}}}   & \multicolumn{1}{c|}{\textcolor{customgreen}{\ding{51}}}     & \multicolumn{1}{c}{\textcolor{customgreen}{\ding{51}}}      &   \textcolor{customred}{\ding{55}}      \\ 
\rowcolor{orange!15} GCD         & \multicolumn{1}{c}{\textcolor{customgreen}{\ding{51}}}   & \multicolumn{1}{c|}{\textcolor{customgreen}{\ding{51}}}     & \multicolumn{1}{c}{\textcolor{customgreen}{\ding{51}}}      &   \textcolor{customgreen}{\ding{51}}   \\ 
\hline
\end{tabular}}
\vspace{-15pt}
\end{table}

%% file: Secs/3_categorization.tex
\section{Category Discovery}
\label{Sec:Methods}
\noindent
\textit{Overview: }
In this section, we present a comprehensive taxonomy of category discovery methods, organized from two {levels: \textit{base settings} and \textit{derived settings}. 
In Sec.~\ref{Subsec:concept}, we introduce two base settings of Category Discovery - NCD and GCD based on how they define the label space for unlabelled data.} In Sec.~\ref{Subsec:Scenario}, we shift focus to the specific objectives and application contexts, grouping them into seven derived settings: CCD, OCD, CD with Domain Shift, DA-CD, SCD, FS-CD, and FCD. Finally, Sec.~\ref{Subsec:beyondImages} explores category discovery on other modalities.

\vspace{-8pt}
\begin{figure}[h!]
\centering
\setlength{\abovecaptionskip}{-4pt}
\includegraphics[width=0.45\textwidth]{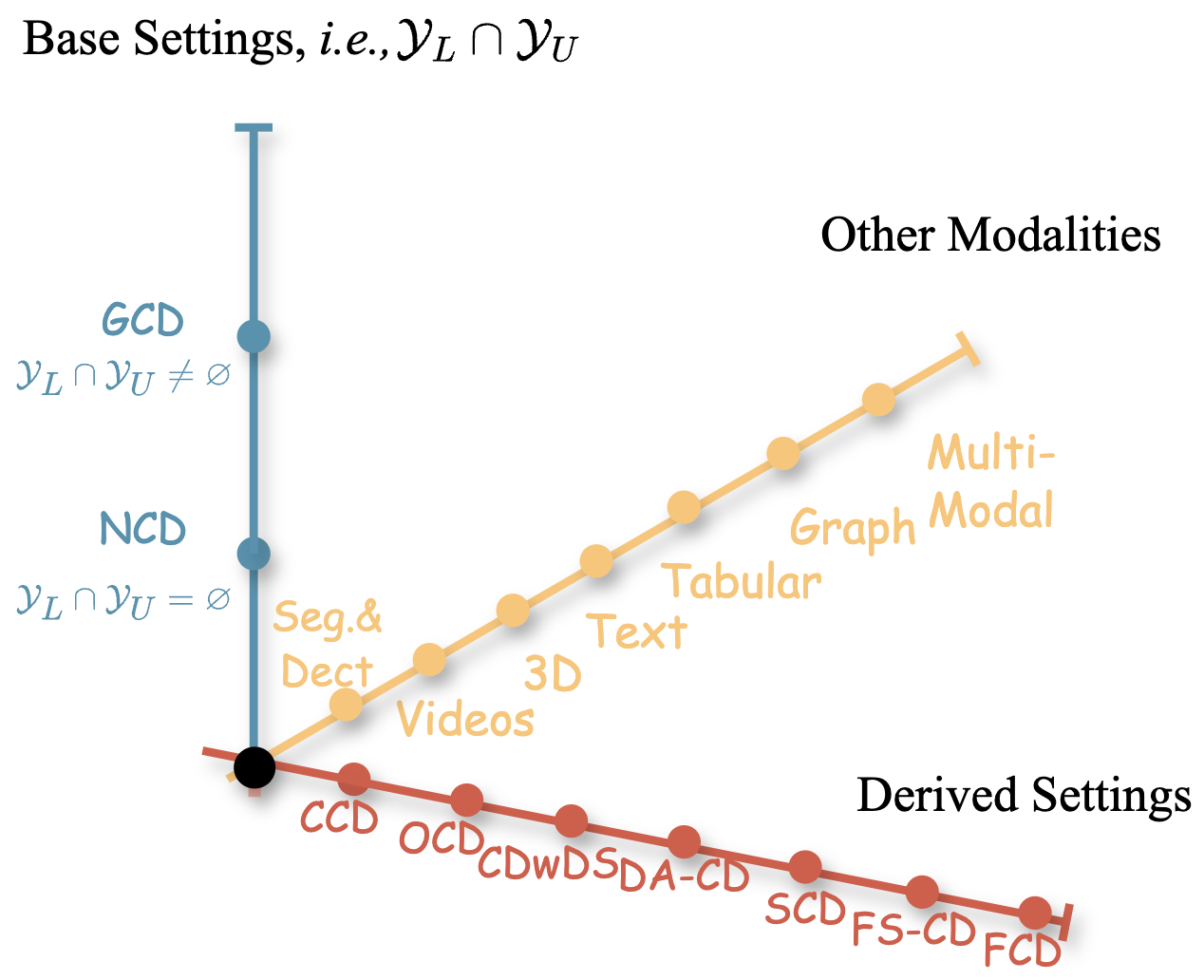}
\put(-200,166.8){{Sec.~\ref{Subsec:concept}}}
\put(-57,130){{Sec.~\ref{Subsec:beyondImages}}}
\put(-50,20.55){{Sec.~\ref{Subsec:Scenario}}}
\caption{ 
{\textbf{Multi-faceted taxonomy.} {We present a multi-faceted taxonomy of Category Discovery spanning base settings, derived settings, and other modalities.} }
}
\label{fig:taxdim}
\vspace{-12pt}
\end{figure}

\hfill \break
\textit{Motivation of Taxonomy:} A single-faceted taxonomy is insufficient to capture the diversity of methodologies in category discovery.
Given the variety of settings, each with its unique challenges and objectives, a more nuanced taxonomy is necessary to categorize these methods effectively. 
Therefore, we offer a comprehensive review of the different configurations of category discovery, considering both \textit{base} and more challenging \textit{derived} settings.

%



\vspace{-10pt}
\subsection{{Base Settings}}
\label{Subsec:concept}

In this subsection, we introduce two base settings based on the definition of the label space for unlabelled data, categorizing methods into NCD and GCD where detailed definitions are provided in Sec.~\ref{Subsec:concept_def}. 
We then discuss conventional category discovery methods for image classification within this taxonomy, while an overview of methods addressing other scenarios and data modalities is presented in Sec.~\ref{Subsec:Scenario} and Sec.~\ref{Subsec:beyondImages}. 
Both NCD and GCD frameworks generally consist of three core components: \textit{representation learning}, \textit{label assignment}, and \textit{estimation of the number of categories}. 
We provide a comprehensive understanding of how they address these key aspects of category discovery.

\input{Parts/NCD}

\input{Parts/GCD_v2}

\vspace{-10pt}
\subsection{Derived Settings}
\label{Subsec:Scenario}
In the following subsection, {we introduce derived settings, which extend category discovery in more practical and challenging tasks to base settings.}
We identify seven distinct derived settings—\textit{CCD}, \textit{OCD}, \textit{CD with Domain Shift}, \textit{DA-CD}, \textit{SCD}, \textit{FS-CD}, and \textit{FCD}. Additionally, we offer an in-depth review of the methods within each setting, highlighting their unique approaches, challenges, and contributions to the category discovery community.

\input{Parts/CCD}

\input{Parts/OCD}

\input{Parts/CDwdomainShift}

\input{Parts/LongTailCD}

\input{Parts/SCD}
\input{Parts/few-shotsCD}
\input{Parts/FCD}

\vspace{-10pt}
\subsection{{Category Discovery on other modalities}}
\label{Subsec:beyondImages}

\input{Parts/OthersCD}


%% file: Parts/NCD.tex
\input{tabs/NCD_table}
\noindent \underline{\textit{\textbf{Novel Category Discovery}.}}
\label{Subsubsec:ncd}
{Han~\etal~\cite{Han2019DTC} introduce Novel Category Discovery (NCD) by adapting Deep Embedded Clustering (DEC)~\cite{dec2016} within a transfer learning framework that integrates representation bottlenecks, temporal ensembling, and consistency regularization to cluster novel categories in unlabelled data.}
{KCL~\cite{Hsu18_L2C} and MCL~\cite{Hsu19_MCL} which are originally designed for cross-task transfer, can also be adopted for NCD. They employ similarity prediction networks trained on labelled data to enhance classification on unlabelled categories.}
%

{Early NCD methods largely focus on adapting clustering and self-supervised learning techniques to discover novel categories, often integrating transfer learning, representation learning, and pseudo-labelling strategies within unified training frameworks. }
RankStats (aka AutoNovel)~\cite{han2019automatically, han21autonovel} leverages self-supervised learning on both labelled and unlabelled data to mitigate bias toward known categories, generating pseudo-labels using rank statistics followed by fine-tuning.
\cite{QING202124} introduces an end-to-end learning framework integrating pairwise similarity learning with self-supervised learning. 
NCL~\cite{ncl} utilises the local neighbourhood of samples in the embedding space to generate pseudo-positive pairs and hard negatives, enhancing contrastive learning. 
OpenMix~\cite{openmix2020} applies MixUp~\cite{mixup} to blend labelled and unlabelled samples, generating more reliable pseudo-labels and uses high-confidence unlabelled samples as reliable anchors to enhance the training process. 
%
%
UNO~\cite{uno} unifies labelled and unlabelled data classification with a multi-view self-labelling strategy, leveraging the Sinkhorn-Knopp algorithm~\cite{Sinkhorn2013} to generate balanced, high-quality pseudo-labels.
DualRS~\cite{dualRs} uses a bi-branch learning framework, combining global and local features with dual ranking statistics and mutual knowledge distillation. 
Squared Mutual Information (SMI)~\cite{9747827} proposes a dependency-based approach that uses SMI to bridge supervised learning and unsupervised clustering. 
\cite{joseph2022spacinglossdiscoveringnovel} introduce Spacing Loss, a simple yet effective method by improving the separability of categories in the latent space. 
ResTune~\cite{ResTune} introduces a disentangled representation learning framework for NCD, separating the learning process into basic and residual features. The framework optimizes the combined representation using three objectives: clustering for novel categories, knowledge distillation to prevent forgetting, and pairwise labelling to preserve structural integrity. 
PSSCNCD~\cite{PSSCNCD} builds a bipartite anchor graph to model data relations and progressively assigns pseudo-labels through self-supervised learning.

{More recent approaches further introduce explicit constraints, semantic alignment objectives, and advanced architectures to improve feature separability and enhance the effectiveness of novel category discovery.}
{Inter-Class and Intra-Class Constraints (IIC)}~\cite{Li2023ncdiic} improves representation learning by incorporating inter-class and intra-class constraints via symmetric Kullback-Leibler divergence. 
%
%
Li~\etal\cite{li2023closerlooknovelclass, li2023supervised} investigate how semantic similarity between labelled and unlabelled datasets impacts NCD performance. 
Their work introduces Sk-Hurt-NCD~\cite{li2023supervised}, which quantifies semantic similarity through the transfer flow metric and proposes pseudo transfer flow to guide the effective use of supervised knowledge in NCD tasks. 
NSCL~\cite{sun2023nscl} models labelled and unlabelled data as graph nodes and introduces a Spectral Contrastive Loss to discover novel classes by minimizing the spectral decomposition of the adjacency matrix.
%
CRKD~\cite{peiyan2023class} proposes a knowledge distillation framework that transfers class relations from known to novel categories, preserving essential semantic information through a learnable regularization process. 
%
Feng~\etal~\cite{wei2023ncdSkin} introduce a framework for identifying unseen skin lesion categories in medical imaging with an uncertainty-aware multi-view cross-pseudo-supervision strategy. 
%
Liu~\etal~\cite{Liu_2024_CVPR} introduce the Region-Aligned Proxy Learning (RAPL) framework, tailored for ultra-fine-grained novel class discovery. It uses channel-wise region alignment to capture local discriminative features and employs semi-supervised proxy learning to model instance–class relationships.
SCKD~\cite{wang2024selfcooperationknowledgedistillationnovel} uses separate representation spaces for known and novel classes, promoting mutual learning through spatial mutual information and self-distillation. 
%
Zhang~\etal~\cite{10328468} propose a two-stage method that first learns features via self-supervised and prototype learning, then trains a classifier using $k$-means pseudo-labels refined by prototypical similarity.
Hasan~\etal~\cite{hasan2023novelcategoriesdiscoveryconstraints} introduce a distribution learning framework that combines Monte Carlo sampling with neuronal activation patterns to align predicted class probabilities with a Multinoulli distribution.



\textit{Label assignment} in NCD is addressed through both \textit{parametric} and \textit{non-parametric} classifiers. 
Non-parametric approaches, such as those in \cite{ResTune, PSSCNCD, sun2023nscl, 9747827, li2023closerlooknovelclass, li2023supervised, Liu_2024_CVPR}, utilise $k$-means based clustering techniques. DTC~\cite{Han2019DTC} employs soft clustering by computing probability distributions based on Euclidean distance between feature vectors and class centres. 
In contrast, many other methods~\cite{Hsu18_L2C, Hsu19_MCL, han2019automatically, QING202124, han21autonovel, openmix2020, ncl, Jia2021JointRL,uno, dualRs, Li2023ncdiic, peiyan2023class, wei2023ncdSkin, wang2024selfcooperationknowledgedistillationnovel, 10328468, hasan2023novelcategoriesdiscoveryconstraints} utilise parametric classifiers, employing classification heads to assign labels to data points.

\textit{Estimating the number of novel categories} in NCD remains a significant challenge, and only a few methods have explicitly tackled this problem. 
DTC~\cite{Han2019DTC} tackles this by semi-supervised $k$-means with probe classes, selecting the optimal count via joint optimization of probe accuracy and clustering validity.
In contrast, most other methods either assume a predefined number of novel categories~\cite{han2021autonovel}, overestimate the category number~\cite{Hsu19_MCL}, or adopt existing approaches, which simplify the process but limit applicability in open-world scenarios.

{
NCD methods demonstrate strong potential for open-world generalization by clustering unlabelled data through knowledge transfer from labelled base categories. 
However, most approaches are trained from scratch, overlooking powerful pretrained features from self-supervised models like DINO~\cite{caron2021dino} and MAE~\cite{MaskedAutoencoders2021}, which may limit representation quality in low-data regimes. 
More critically, NCD assumes that unlabelled data consists only of novel categories—a convenient but unrealistic setup. In practice, unlabelled sets often mix seen and unseen classes, motivating the shift toward the more practical Generalized Category Discovery.
}

%% file: tabs/NCD_table.tex
\begin{table*}[th]
\centering
\caption{Summary of essential components for reviewed NCD approaches.}
\vspace{-6pt}
\label{table:NCD}
\begin{threeparttable}
	\resizebox{0.75\textwidth}{!}{
		\setlength\tabcolsep{6pt}
		\renewcommand\arraystretch{1.0}
		\begin{tabular}{|c|r||c|c|c|}
			\hline
			\rowcolor[HTML]{D3D3D3}
			Year & Method~~~~ & Pub. & Backbone & Label Assignment \\
			\hline
			\hline
            \hline
			\multirow{1}{*}{{2018}}
            & KCL~\cite{Hsu18_L2C} & \textit{ICLR} & ResNet & Parametric Classifier \\
            \hline
			\multirow{2}{*}{{2019}}
            & MCL~\cite{Hsu19_MCL} & \textit{ICLR} & ResNet, VGG, LeNet & Parametric Classifier \\
            & DTC~\cite{Han2019DTC} & \textit{ICCV} & ResNet, VGG & Soft Assignment \\
            \hline
			\multirow{1}{*}{{2020}}
            & RankStats,RankStats+~\cite{han2019automatically} & \textit{ICLR} & ResNet & Parametric Classifier \\
            \hline
			\multirow{7}{*}{{2021}}
            & Qing \etal~\cite{QING202124} & \textit{Neural Networks} & ResNet & Parametric Classifier \\
            & OpenMix~\cite{openmix2020} & \textit{CVPR} & ResNet, VGG & Parametric Classifier \\
            & NCL~\cite{ncl} & \textit{CVPR} & ResNet & Parametric Classifier \\
            & UNO~\cite{uno} & \textit{ICCV} & ResNet & Parametric Classifier \\
            & DualRS~\cite{dualRs} & \textit{ICCV} & ResNet & Parametric Classifier \\
            \hline
            \multirow{3}{*}{{2022}}
            & SMI~\cite{9747827} & \textit{ICASSP} & VGG & $k$-means \\
            & PSSCNNCD~\cite{PSSCNCD} & \textit{T-CYB} & N/A & BKBH $k$-means \\
            & Li \etal~\cite{li2023closerlooknovelclass} & \textit{NeurIPSW} & ResNet & $k$-means \\
            \hline
            \multirow{6}{*}{{2023}}
            & ResTune~\cite{ResTune} & \textit{T-NNLS} & ResNet & $k$-means \\
            & SK-Hurt~\cite{li2023supervised} & \textit{TMLR} & ResNet & $k$-means \\
            & IIC~\cite{Li2023ncdiic} & \textit{CVPR} & ResNet & Parametric Classifier \\
            & NSCL~\cite{sun2023nscl} & \textit{ICML} & ResNet & $k$-means \\
            & CRKD~\cite{peiyan2023class} & \textit{ICCV} & ResNet, ViT & Parametric Classifier \\
            & Feng \etal\cite{wei2023ncdSkin} & \textit{MICCAI} & ResNet & Parametric Classifier \\
            \hline
			\multirow{3}{*}{{2024}}
            & RAPL~\cite{Liu_2024_CVPR} & \textit{CVPR} & ResNet & $k$-means \\
            & SCKD~\cite{wang2024selfcooperationknowledgedistillationnovel} & \textit{ECCV} & ResNet,ViT & Parametric Classifier \\
            & APL~\cite{10328468} & \textit{T-PAMI} & ResNet & Parametric Classifier \\
            \hline
			\multirow{1}{*}{{PrePrint}}
            & Hasan \etal~\cite{hasan2023novelcategoriesdiscoveryconstraints} & \textit{ArXiv} & ResNet & Parametric Classifier \\
            \hline
		\end{tabular}
        }
\end{threeparttable}
\vspace{-12pt}
\end{table*}

%% file: Parts/GCD_v2.tex
\input{tabs/gcd_table}
\noindent \underline{\textit{\textbf{Generalized Category Discovery}.}}
\label{Subsecsec:gcd}
%
Vaze~\etal~\cite{vaze2022gcd} extend NCD~\cite{Han2019DTC} to GCD by introducing a decoupled representation learning approach. 
Unlike NCD methods, which mainly train models from scratch, they use a large-scale self-supervised pre-trained model (\eg, DINO~\cite{caron2021dino}) for initialization. 
The model is then fine-tuned with both supervised and unsupervised contrastive learning on labelled and unlabelled data. 
%
ComEx~\cite{yang2022comex} introduces batch-wise and class-wise experts to manage both base and novel classes, learning discriminative representations across them.
NACH~\cite{guo2022nach} mitigates the learning rate disparity between seen and unseen classes by synchronizing learning through pairwise similarity and adaptive distribution alignment.
XCon~\cite{fei2022xcon} partitions the dataset into sub-datasets using $k$-means clustering based on DINO~\cite{caron2021dino} representations, followed by contrastive learning to capture fine-grained features.

{Recent advances in GCD further explore prototype-driven, prompt-based, hierarchical, and hash-based designs, offering complementary strengths to contrastive learning-based baselines and facilitating more effective category discovery.}
PromptCAL~\cite{zhang2022promptcal} introduces a two-stage prompt-based framework, combining discriminative prompt regularization and contrastive affinity learning to iteratively refine semantic representations. 
%
%
DCCL~\cite{pu2023dynamic} proposes dynamic conceptual contrastive learning, alternating between instance-level and abstract conception-level with dynamically generated conceptions.
GPC~\cite{Zhao_2023_ICCV} uses a semi-supervised Gaussian Mixture Model (GMM) with dynamic splitting and merging, along with Prototypical Contrastive Learning, to improve feature representation.
PIM~\cite{chiaroni2023parametric} introduces bi-level optimization to maximize mutual information between data features and labels while addressing class-balance bias with a parametric family of objectives.
%
$\mu$GCD~\cite{vaze2023clevr4} introduces the Clevr-4 dataset, designed to evaluate GCD methods across different taxonomies, and highlights potential biases in large pre-trained models like DINO~\cite{caron2021dino} and MAE~\cite{MaskedAutoencoders2021}. {It adopts a mean-teacher approach to overcome GCD limitations of unstable pseudo-labels.}
InfoSieve~\cite{rastegar2023learn} generates binary category codes through self-supervised contrastive learning and optimizes mutual information between features and codes, organizing them in a hierarchical binary tree for efficient categorization.
Extended from NSCL~\cite{sun2023nscl}, SORL~\cite{sun2024graph} introduces a graph-theoretic framework that uses spectral decomposition to improve data representation by incorporating both self-supervised and supervised signals.
%
CAC~\cite{yang2023GCDwithclustering} employs weak and strong augmentations to generate diverse views of samples, ensuring clustering assignment consistency via feature-prototype similarity alignment.
{CiPR~\cite{hao2023cipr} introduces cross-instance positive relations between labelled and unlabelled data by leveraging a novel selective neighbor clustering algorithm, which generates high-purity pseudo-labels to strengthen representation learning.}
CMS~\cite{choi2024contrastive} combines contrastive learning with mean-shift clustering to refine image embeddings by generating mean-shifted representations through nearest neighbors.
Yang~\etal~\cite{yang2024learning} propose a Neighbor Graph Convolutional Network for improved pseudo-labelling by leveraging $k$-nearest neighbor relationships and iterative contrastive learning at both instance and class levels.
SelEx~\cite{RastegarECCV2024} leverages hierarchical semi-supervised $k$-means pseudo-labelling and self-expertise to refine clustering and adapt contrastive learning, enhancing fine-grained classification and novel category generalization.
ConceptGCD~\cite{zhang2024composing} employs a three-stage approach that learns known class concepts, composes derivable novel concepts, and discovers underivable ones via contrastive learning and normalization.
PNP~\cite{wang2024knownclustersprobenew} integrates learnable potential prototypes with self-distillation to refine cluster centers in an end-to-end manner.
RPIM~\cite{tan2024revisitingmutualinformationmaximization} extends PIM~\cite{chiaroni2023parametric} with regularization to address unconfident predictions and feature refinement using semantic-bias transformations.
MSGCD~\cite{DUAN2025103020} unifies parametric classification and representation learning using a mutual-support mechanism, enhancing both pseudo-labels and feature representations iteratively.
NN-GCD~\cite{ji2024fresh} reformulates GCD as an optimization problem, proving that optimal $k$-means clustering can be achieved using Symmetric Non-negative Matrix Factorization and Non-negative Contrastive Learning.
DIG-FACE~\cite{luo2024digfacedebiasedlearninggeneralized} mitigates implicit bias via F-discrepancy and applies explicit debiasing to enhance facial expression recognition in GCD.
{
Dai~\etal~\cite{dai2025adaptive} introduces a plug-and-play adaptive part learning module that uses DINO-derived part priors and shared learnable queries to discover consistent object parts without annotations.
HypCD~\cite{liu2025hyperbolic} pioneers the use of hyperbolic embeddings to capture hierarchical class relations and introduces a general GCD framework operating in hyperbolic space.}
ConGCD~\cite{tang2025dissecting} deconstructs an image into competing visual primitives via reconstruction-guided slot attention, then injects a Multiplex Consensus module into the ViT FFN to aggregate them. 
AF~\cite{xu2025hidden} proposes Attention Focusing, a plug-and-play module that combats distracted attention in GCD by learning token importance from labelled data and adaptively pruning low-value image tokens.

SimGCD~\cite{wen2023simgcd} introduces a popular end-to-end parametric baseline that jointly trains a feature extractor and a parametric classifier with self-distillation and entropy regularization, outperforming previous two-stage methods.
{It has since become a popular foundation for developing more advanced GCD approaches. Numerous subsequent works extend or build upon its design to address specific limitations and enhance performance.}
{Building on SimGCD~\cite{wen2023simgcd}, SPT-Net~\cite{wang2024sptnet} presents a two-stage iterative framework introducing additional spatial prompt tuning, optimizing image patch-based discriminative features.}
DebGCD~\cite{liu2025dg} introduces an auxiliary debiased classifier to SimGCD~\cite{wen2023simgcd} trained with hard labels to mitigate label bias, alongside a semantic distribution detector that identifies shifts between known and unknown classes in a separate feature space.
%
AMEND~\cite{Banerjee_2024_WACV} proposes Expanded Neighbourhood Contrastive Learning using both nearest and expanded neighbors for robust contrastive pairing, along with a Class-wise Adaptive Margin Regularizer to improve fine-grained category separation.
LegoGCD~\cite{Cao_2024_CVPR} builds on SimGCD to mitigate catastrophic forgetting by introducing Local Entropy Regularization for stabilizing high-confidence predictions and Dual-view KL Divergence to improve reliable known-sample selection.
FlipClass~\cite{lin2024flipped} proposes dynamic teacher-student attention alignment, based on Hopfield Network energy functions, by updating the teacher's attention using student feedback through energy minimization.
Contextuality-GCD~\cite{luo2024contextualityhelpsrepresentationlearning} augments contrastive learning with instance-level context and cluster-level prototypical cues.
PAL-GCD~\cite{wang2025prior} introduces a greedy association algorithm for Prior-Constrained Association Learning to SimGCD~\cite{wen2023simgcd}, ensuring each cluster contains at most one known category to avoid incorrect merging.
ProtoGCD~\cite{protoGCD} proposes a unified prototype-based framework that introduces a shared prototype classifier and a dual-level adaptive pseudo-labelling mechanism to mitigate confirmation bias and balance learning across base and novel classes. 
MCDL~\cite{tu2024memoryconsistencyguideddivideandconquer} uses dual memory banks to assess sample reliability and splits data by intra- and inter-memory consistency for targeted learning.
ActiveGCD~\cite{ma2024active} introduces an active-learning framework to select informative samples from unlabelled data, ensuring robust label assignment through an exponential moving average model.
{
MOS~\cite{peng2025mos} employs an extra zero-shot saliency segmentation model to decouple object and scene regions, and feeds both through a shared backbone with an MLP-based scene-awareness module to model object–scene associations.
AptGCD~\cite{zhang2025less} leverages inner-patch Meta Visual Prompts and a Prompt Transformer with global self-attention to efficiently fuse local and global features for distinguishing base and novel classes.}
{SEAL~\cite{he2025seal} introduces a semantic-aware hierarchical learning framework for GCD, where semantically consistent coarse-to-fine supervision is propagated from labelled to unlabelled data via semantic-aware hierarchical contrastive learning and a cross-granularity consistency module.}

{In parallel with the development of GCD, a number of methods originating from the Open-World Semi-Supervised Learning (OWSSL) community—conceptually equivalent to GCD—have been proposed to address the challenge of discovering novel categories under partially labelled data. 
}
~\cite{orca2022} introduces ORCA, an end-to-end framework that adaptively adjusts decision margins based on uncertainty, enabling robust learning across both seen and unseen categories. 
Building upon this foundation, OwMatch~\cite{niu2024owmatch} enhances self-labelling by conditioning pseudo-labels on labelled data to reduce confirmation bias, and introduces hierarchical thresholding to balance confidence between base and novel classes.
Similarly, LPS\cite{lps_ijcai2024} focuses on addressing learning imbalance by combining an adaptive margin loss with a contrastive clustering objective tailored for novel class discovery.
OpenLDN~\cite{rizve2022openldn} converts OWSSL into a closed-world task by using regression-based pairwise similarity to generate pseudo-labels and integrate novel samples into the labeled set.
OpenCon~\cite{sun2023opencon} utilizes prototype-based learning to generate compact and distinguishable representations, applying contrastive loss to both known and novel data within an OWSSL framework.
TRSSL~\cite{rizve2022towards} introduces a pseudo-labelling method that integrates prior class distribution knowledge and sample uncertainty to generate reliable pseudo-labels for both novel and base classes.
OpenNCD~\cite{ijcai2023p445} utilizes a bi-level contrastive learning approach at both the prototype and group levels to enhance representation learning.
TIDA~\cite{wang2023discover} introduces a hierarchical approach that constructs prototypes at multiple levels (sub-class, target-class, super-class) and aligns predictions across these levels using an affinity matrix.
TRAILER~\cite{Xiao_2024_CVPR} aligns representations to a fixed Neural Collapse-inspired classifier and refines pseudo-labels via hierarchical sample-target allocation using optimal transport and PU learning.
OpenGCD~\cite{gao2023opengcdassistingopenworld} integrates uncertainty-based OSR with GCD to identify unseen categories, and uses semi-supervised $k$-means++ for automated class discovery.


{Another important direction integrates cross-modal knowledge into GCD, with methods leveraging Vision-Language Models (VLMs) and Large Language Models (LLMs) to incorporate textual information, enhancing category discovery for both base and novel classes.}
CLIP-GCD~\cite{ouldnoughi2023clipgcdsimplelanguageguided} is the first to leverage pre-trained VLMs for category discovery by utilizing CLIP's~\cite{clip} multi-modal capabilities. It integrates visual features with relevant textual descriptions from a large corpus, enabling joint optimization of image and text embeddings to capture richer semantic information with auxiliary textual cues.
TextGCD~\cite{zheng2024textualknowledgematterscrossmodality} advances cross-modal learning with a dynamic co-teaching framework, aligning textual and visual features using large language models like GPT-3~\cite{brown2020language}.
%
Additionally, GET~\cite{wang2024unlockingmultimodalpotentialclip} enhances representation learning by integrating visual and textual features without labelled text, utilizing CLIP’s vision-language alignment to generate pseudo-text embeddings.
CPT~\cite{yang2025consistent} adapts CLIP to learn ``task + class'' prompts, ensuring robust category discovery with Vision-Vision and Vision-Language consistency across labelled and unlabelled data.

As with NCD, label assignment plays a critical role in GCD pipelines. Both \textit{parametric} and \textit{non-parametric} classifiers have been explored to support flexible and scalable clustering of novel categories.
GCA~\cite{Otholt_2024_WACV} offers a plug-and-play parametric-classifier solution that seamlessly integrates with existing GCD methods (\eg, \cite{vaze2022gcd} and \cite{zhang2022promptcal}), outperforming traditional clustering methods such as semi-supervised $k$-means, especially in fine-grained tasks. It employs Guided Cluster Aggregation to hierarchically merge small pure clusters into target categories based on local feature structures. 
Among non-parametric approaches, OpenCon~\cite{sun2023opencon} uses prototype-based assignment, selecting the prototype with the highest softmax-normalized similarity to the projected features. GPC~\cite{Zhao_2023_ICCV} applies a semi-supervised Gaussian Mixture Model (GMM) for clustering. 
CMS~\cite{choi2024contrastive} uses the mean-shift algorithm~\cite{meanshift} to iteratively shift data points toward the mode of their neighbours in the embedding space. 
SMILE~\cite{du2023on} employs hash-code-based clustering, while PHE~\cite{zheng2024prototypical} adopts a Hamming ball-based strategy for inference. Both PNP~\cite{wang2024knownclustersprobenew} and DCCL~\cite{pu2023dynamic} utilize Infomap~\cite{rosvall2009map} for clustering, and CAC~\cite{yang2023GCDwithclustering} applies the Louvain~\cite{Louvain} community detection algorithm for label assignment. 
Yang~\etal~\cite{yang2024learning} utilize FINCH~\cite{sarfraz2019finch} and introduces Cross-View Consistency Strategy to refine by comparing clustering results from FINCH~\cite{sarfraz2019finch} and $k$-means. 
CiPR~\cite{hao2023cipr} utilizes its Selective Neighbor Clustering (SNC), a semi-supervised variant of FINCH~\cite{sarfraz2019finch}, to produce hierarchical partitions.
PAL-GCD~\cite{wang2025prior} adopts a two-stage framework, using non-parametric clustering for warm-up, then jointly training with a parametric classifier.
%
%


{Estimating the class number of novel categories remains a core challenge in GCD.}
%
GCD~\cite{vaze2022gcd} addresses this by running $k$-means with different cluster counts and selecting the optimal number via Brent's algorithm to maximize accuracy on labelled data.
OpenNCD~\cite{ijcai2023p445} uses Jaccard-based prototype grouping for class estimation and benchmarks it against X-means~\cite{pmlr-v119-guo20i}, G-means~\cite{hamerly2003learning}, and DipDeck~\cite{leiber2021dip}. 
GPC~\cite{Zhao_2023_ICCV} integrates class estimation with representation learning using a semi-supervised GMM featuring a dynamic cluster splitting and merging mechanism. 
CMS~\cite{choi2024contrastive} applies agglomerative clustering with a Ward linkage criterion to determine optimal clusters during training. 
CiPR~\cite{hao2023cipr} estimates the class number by generating hierarchical partitions through SNC and selecting the optimal level.
CAC~\cite{yang2023GCDwithclustering} utilizes the Louvain~\cite{Louvain} algorithm for class estimation. 
PHE enables on-the-fly class discovery by assigning samples to existing categories if their hash lies within a Hamming ball of known centers, or creating a new category otherwise.
SMILE~\cite{du2023on} also estimates the number of classes with hash-coding by assigning labels to unique binary hashes and counting them as the class count. 
DCCL~\cite{pu2023dynamic} and PNP~\cite{wang2024knownclustersprobenew} leverage Infomap~\cite{rosvall2009map} to detect communities and refine small clusters, determining the final class count. 
{The benchmark for class-number estimation is detailed in the Appendix.}

{
GCD extends NCD to a more realistic scenario where unlabelled data contains both seen and unseen categories, better aligning with real-world conditions and demonstrating the feasibility of open-world learning under partial supervision. 
Yet, many methods remain biased toward seen classes, limiting novel class discovery, and most studies focus narrowly on general image datasets, leaving domains like medical imaging underexplored.
In addition, current approaches often assume balanced distributions and homogeneous domains, which rarely hold in practice. These limitations naturally motivate the development of more complex derived settings.
}

%% file: tabs/gcd_table.tex
\begin{table*}
\centering
\caption{Summary of essential components for reviewed GCD approaches.}
\vspace{-5pt}
\label{table:GCD}
\begin{threeparttable}
	\resizebox{0.65\textwidth}{!}{
		\setlength\tabcolsep{6pt}
		\renewcommand\arraystretch{1.0}
		\begin{tabular}{|c|r||c|c|c|}
			\hline
			\rowcolor[HTML]{D3D3D3}
			Year &Method~~~~  &Pub. &Backbone  & Label Assignment \\
			\hline
			\hline
            \hline
			\multirow{7}{*}{\rotatebox{90}{2022}}
            &GCD~\cite{vaze2022gcd}&\textit{CVPR}&ViT&$k$-means\\
            &ORCA~\cite{orca2022}&\textit{CVPR}&ResNet&Parametric Classifier\\
            &ComEx~\cite{yang2022comex}&\textit{CVPR}&ResNet&Parametric Classifier\\
            &OpenLDN~\cite{rizve2022openldn}&\textit{ECCV}&ResNet&Parametric Classifier\\
            &TRSSL~\cite{rizve2022towards}&\textit{ECCV}&ResNet&Parametric Classifier\\
            &NACH~\cite{guo2022nach}&\textit{NeurIPS}&ResNet&Parametric Classifier\\
            &XCon~\cite{fei2022xcon}&\textit{BMVC}&ViT&$k$-means\\
            \hline
			\multirow{13}{*}{\rotatebox{90}{2023}}
            &OpenCon~\cite{sun2023opencon}&\textit{TMLR}&ResNet&Prototype-based\\
            &PromptCAL~\cite{zhang2022promptcal}&\textit{CVPR}&ViT&$k$-means\\
            &DCCL~\cite{pu2023dynamic}&\textit{CVPR}&ViT&Infomap\\
            &OpenNCD~\cite{ijcai2023p445}&\textit{IJCAI}&ResNet&Prototype-based\\
            &SimGCD~\cite{wen2023simgcd}&\textit{ICCV}&ViT&Parametric Classifier\\
            &GPC~\cite{Zhao_2023_ICCV}&\textit{ICCV}&ViT&GMM\\
            &PIM~\cite{chiaroni2023parametric}&\textit{ICCV}&ViT&Parametric Classifier\\
            &TIDA~\cite{wang2023discover}&\textit{NeurIPS}&ResNet&Parametric Classifier\\
            &$\mu$GCD~\cite{vaze2023clevr4}&\textit{NeurIPS}&ResNet, ViT&$k$-means\\
            &InfoSieve~\cite{rastegar2023learn}&\textit{NeurIPS}&ViT&$k$-means\\
            &SORL~\cite{sun2024graph}&\textit{NeurIPS}&ResNet&$k$-means\\
            &CAC~\cite{yang2023GCDwithclustering}&\textit{ICONIP}&ViT&Louvain\\
            \hline
			\multirow{15}{*}{\rotatebox{90}{2024}}
            &CiPR~\cite{hao2023cipr}&\textit{TMLR}&ViT&\textit{SNC}\\
            &AMEND~\cite{Banerjee_2024_WACV}&\textit{WACV}&ViT&Parametric Classifier\\
            &GCA~\cite{Otholt_2024_WACV}&\textit{WACV}&ViT&Guided Cluster Aggregation\\
            &SPT-Net~\cite{wang2024sptnet}&\textit{ICLR}&ViT&Parametric Classifier\\
            &LegoGCD~\cite{Cao_2024_CVPR}&\textit{CVPR}&ViT&Parametric Classifier\\
            &CMS~\cite{choi2024contrastive}&\textit{CVPR}&ViT&Agglomerative Clustering\\
            &TRAILER~\cite{Xiao_2024_CVPR}&\textit{CVPR}&ResNet, ViT&Parametric Classifier\\
            &ActiveGCD~\cite{ma2024active}&\textit{CVPR}&ViT&Parametric Classifier\\
            &TextGCD~\cite{zheng2024textualknowledgematterscrossmodality}&\textit{ECCV}&ViT&CVCS\\
            &Yang~\etal~\cite{yang2024learning}&\textit{ECCV}&ViT&Parametric Classifier\\
            &SelEx~\cite{RastegarECCV2024}&\textit{ECCV}&ViT&$k$-means\\
            &LPS~\cite{lps_ijcai2024}&\textit{IJCAI}&ResNet&Parametric Classifier\\
            &OwMatch~\cite{niu2024owmatch}&\textit{NeurIPS}&ResNet&Parametric Classifier\\
            &FlipClass~\cite{lin2024flipped}&\textit{NeurIPS}&ViT&Parametric Classifier\\
            &Contextuality-GCD~\cite{luo2024contextualityhelpsrepresentationlearning}&\textit{ICIP}&ViT&Parametric Classifier\\
            \hline
			\multirow{12}{*}{\rotatebox{90}{2025}}
            &MSGCD~\cite{DUAN2025103020}&\textit{Information Fusion}&ViT&Parametric Classifier\\
            &CPT~\cite{yang2025consistent}&\textit{IJCV}&ViT&Similarity-based\\
            &PAL-GCD~\cite{wang2025prior}&\textit{AAAI}&ViT&Parametric Classifier\\
            &DebGCD~\cite{liu2025dg}&\textit{ICLR}&ViT&Parametric Classifier\\
            &ProtoGCD~\cite{protoGCD}&\textit{T-PAMI}&ViT&Parametric Classifier\\
            &MOS~\cite{peng2025mos}&\textit{CVPR}&ViT&Parametric Classifier\\
            &GET~\cite{wang2024unlockingmultimodalpotentialclip}&\textit{CVPR}&ViT&Parametric Classifier\\
            &AptGCD~\cite{zhang2025less}&\textit{CVPR}&ViT&Parametric Classifier\\
            &Dai~\etal~\cite{dai2025adaptive}&\textit{CVPR}&ViT&-\\
            &HypCD~\cite{liu2025hyperbolic}&\textit{CVPR}&ViT&-\\
            &ConGCD~\cite{tang2025dissecting}&\textit{ICCV}&ViT&-\\
            &AF~\cite{xu2025hidden}&\textit{ICCV}&ViT&-\\
            &SEAL~\cite{he2025seal}&\textit{NeurIPS}&ViT&Parametric Classifier\\
            \hline
			\multirow{8}{*}{\rotatebox{90}{PrePrint}}
            &CLIP-GCD~\cite{ouldnoughi2023clipgcdsimplelanguageguided}&\textit{ArXiv}&ViT&$k$-means\\
            &MCDL~\cite{tu2024memoryconsistencyguideddivideandconquer}&\textit{ArXiv}&ViT&Parametric Classifier\\
            &PNP~\cite{wang2024knownclustersprobenew}&\textit{ArXiv}&ViT&Infomap\\
            &RPIM~\cite{tan2024revisitingmutualinformationmaximization}&\textit{ArXiv}&ViT&Parametric Classifier\\
            &OpenGCD~\cite{gao2023opengcdassistingopenworld}&\textit{ArXiv}&ViT&Parametric Classifier\\
            &ConceptGCD~\cite{zhang2024composing}&\textit{ArXiv}&ViT&Parametric Classifier\\
            &DIG-FACE~\cite{luo2024digfacedebiasedlearninggeneralized}&\textit{ArXiv}&ViT&Parametric Classifier\\
            &NN-GCD~\cite{ji2024fresh}&\textit{ArXiv}&ViT&SNMF\\
            \hline
		\end{tabular}
        }
\end{threeparttable}
\vspace{-15pt}
\end{table*}

%% file: Parts/CCD.tex
\input{tabs/CCD_table}

\noindent \underline{\textit{\textbf{Continual Category Discovery (CCD)}.}}
\label{Subsecsec:ccd}
CCD~\cite{joseph2022novelclassdiscoveryforgetting} provides a continual setting of category discovery in which new categories are identified sequentially while retaining previously acquired knowledge.
It begins with an initial stage, where the model is trained on a labelled dataset, denoted as $\mathcal{D}_{train}^0 = \{(\bm{x_i^0},y^0_i)\}^{N^0}_{i=1}$, composed of instances from base categories $\mathcal{C}_B^0$. Subsequently, the model enters a discovery phase in which it is incrementally exposed to new datasets $\{\mathcal{D}_{train}^t\}_{t=1}^T$ over a series of timesteps $\{1, \ldots, T\}$. 
At each timestep, the model is retrained using the current dataset $\mathcal{D}_{train}^t$ to identify novel categories while preserving previously acquired knowledge.

CCD presents several distinct scenarios based on the structure of the incoming data. 
In the \emph{Class Incremental Scenario}~\cite{joseph2022novelclassdiscoveryforgetting, roy2022class, zhang2022growmergeunifiedframework, liu2023large}, the training set $\mathcal{D}_{train}^t$ contains solely unlabelled instances from novel categories.
%
NCDwF~\cite{joseph2022novelclassdiscoveryforgetting} generates pseudo-latent representations to serve as proxies for the discarded labelled data and employs a mutual-information-based regularizer to both enhance novel class discovery and retain performance on known classes. 
%
%
Similarly, FRoST~\cite{roy2022class} prevents forgetting by storing feature prototypes and applying feature-level knowledge distillation, while Msc-iNCD ~\cite{liu2023large} leverages self-supervised pre-trained models like DINO~\cite{caron2021dino} along with cosine normalization and feature replay to mitigate catastrophic forgetting. 
ADM~\cite{chen2024adaptivediscoveringmergingincremental} decouples representation learning from category discovery using Triplet Comparison and Probability Regularization, and employs Adaptive Model Merging to integrate new knowledge without increasing model size or causing forgetting.
FEA further enhances features using a single-head paradigm guided by a prior distribution, incorporating Centroid-to-Samples Similarity and Boundary-Aware Prototype constraints to maintain clear boundaries between known and novel classes.

\emph{Class Incremental Scenario} assumes that each incoming dataset at a timestep contains only novel categories, which is often an unrealistic premise. In contrast, \emph{mixed-incremental scenario} assumes that $\mathcal{D}_{train}^t$ includes unlabelled data from both base and novel categories, as seen in works like \cite{Kim_2020_pa, wu2023metagcd, park2024onlinecontinuousgeneralizedcategory, cendra2024promptccd}.
%
PA-GCD~\cite{Kim_2023_pacgcd} uses a proxy-anchor approach with affinity propagation~\cite{Frey_Dueck_2007} for clustering and class estimation, removing the need to predefine category numbers. It also leverages metric-based exemplars to reduce forgetting.
MetaGCD~\cite{wu2023metagcd} presents a meta-learning framework that aligns the training objective with the evaluation metric to prevent forgetting and enhances feature learning via soft neighborhood contrastive learning with adaptive positive sample selection.
PromptCCD~\cite{cendra2024promptccd} introduces a Gaussian Mixture Prompting module to dynamically manage prompts for continual category discovery. It combines contrastive learning with GMM-based prompt retrieval to support on-the-fly category estimation and knowledge retention. Class number is estimated via a GMM with a split-and-merge strategy guided by the Hastings ratio.
DEAN~\cite{park2024onlinecontinuousgeneralizedcategory} separates seen and unseen samples via energy-guided discovery, enhances clustering with variance-based augmentation, and applies energy-based contrastive loss to improve novel class discrimination while mitigating forgetting. Class number is estimated via affinity propagation.
Happy~\cite{ma2024happy} mitigates prediction and hardness biases via clustering-guided initialization and entropy regularization, and uses Hardness-Aware Prototype Sampling to model class distributions. Class number is estimated by maximizing the silhouette score.
{VB-CGCD~\cite{dai2025vbcgcd} formulates continual category discovery from a Bayesian perspective, modelling each class as a variational Gaussian distribution with covariance-aware nearest-class-mean classification.}

\emph{Self-supervised mixed-incremental scenario} assumes that $\mathcal{D}_{train}^t$ includes both labelled and unlabelled data from base and novel categories. 
Under this assumption, iGCD~\cite{zhao2023incremental} combines a non-parametric Soft Nearest-Neighbor classifier with density-based sampling to classify both known and novel categories. It selects density peaks as support samples to reduce forgetting and estimates the class number by counting these peaks.
CAMP~\cite{rypesc2024category} proposes a three-phase framework combining contrastive learning, semi-supervised clustering, and centroid adaptation to balance novel class discovery and knowledge retention.

Additionally, GM~\cite{zhang2022growmergeunifiedframework} evaluates all these scenarios by introducing an inductive continual category discovery setting. In this approach, the model is first trained on a labelled dataset and then updated across successive datasets, with testing on unlabelled sets at each timestep. It alternates between learning novel categories and merging them into a static branch via distillation, enabling continual discovery while preserving prior knowledge.

%% file: tabs/CCD_table.tex
\begin{table*}[ht]
\centering
\caption{Summary of essential components for reviewed CCD approaches.}
\vspace{-6pt}
\label{table:CCD}
\begin{threeparttable}
	\resizebox{0.9\linewidth}{!}{
		\setlength\tabcolsep{6pt}
		\renewcommand\arraystretch{1.0}
		\begin{tabular}{|c|r||c|c|c|c|c|}
			\hline
			\rowcolor[HTML]{D3D3D3}
			Year & Method~~~~ & Pub. & Backbone & Scenario & Label Assignment & $\mathcal{Y_L} \cap \mathcal{Y_U}$\\
			\hline
			\hline
			\hline
			\multirow{3}{*}{\rotatebox{90}{2022}}
			& NCDwF~\cite{joseph2022novelclassdiscoveryforgetting} & \textit{ECCV} & ResNet & Class Incremental & Parametric Classifier & $\varnothing$\\
            & FRoST~\cite{roy2022class} & \textit{ECCV} & ResNet & Class Incremental & Parametric Classifier & $\varnothing$\\
            & GM~\cite{zhang2022growmergeunifiedframework} & \textit{NeurIPS} & ResNet & All & Parametric Classifier & $ \neq \varnothing$\\
            \hline
			\multirow{3}{*}{\rotatebox{90}{2023}}
            & PA-GCD~\cite{Kim_2023_pacgcd} & \textit{ICCV} & ViT,ResNet & Mixed Incremental & Parametric Classifier & $ \neq \varnothing$\\
            & MetaGCD~\cite{wu2023metagcd} & \textit{ICCV} & ViT & Mixed Incremental & $k$-means & $ \neq \varnothing$\\
            & iGCD~\cite{zhao2023incremental} & \textit{ICCV} & ResNet & Self-Supervised Mixed Incremental & Soft Nearest Neighbor & $ \neq \varnothing$\\
            \hline
			\multirow{6}{*}{\rotatebox{90}{2024}}
            & Msc-iNCD~\cite{liu2023large} & \textit{ICPR} & ViT & Class Incremental & Parametric Classifier & $\varnothing$\\
            & ADM~\cite{chen2024adaptivediscoveringmergingincremental} & \textit{AAAI} & ResNet & Class Incremental & Parametric Classifier & $\varnothing$\\
            & PromptCCD~\cite{cendra2024promptccd} & \textit{ECCV} & ViT & Mixed Incremental & GMM & $ \neq \varnothing$\\
            & DEAN~\cite{park2024onlinecontinuousgeneralizedcategory} & \textit{ECCV} & ViT & Mixed Incremental & Parametric Classifier & $ \neq \varnothing$\\
            & CAMP~\cite{rypesc2024category} & \textit{ECCV} & ViT & Self-Supervised Mixed Incremental & Nearest Centroid Classifier & $ \neq \varnothing$\\
            & Happy~\cite{ma2024happy} & \textit{NeurIPS} & ViT & Mixed Incremental & Parametric Classifier & $ \neq \varnothing$\\
            \hline
			\multirow{1}{*}{{2025}}
            & VB-CGCD~\cite{dai2025vbcgcd} & \textit{ICML} & ViT & Mixed Class Incremental & Variational Bayes NCM & $\neq \varnothing$\\
            \hline
			\multirow{1}{*}{{preprint}}
            & FEA~\cite{yu2024continualnovelclassdiscovery} & \textit{ArXiv} & ViT & Class Incremental & Parametric Classifier & $\varnothing$\\
            \hline
		\end{tabular}
        }
\end{threeparttable}
\vspace{-15pt}
\end{table*}

%% file: Parts/OCD.tex
\input{tabs/OCD_table}

\input{tabs/CDwDomain_shift_table}

\noindent \underline{\textit{\textbf{On-the-fly Category Discovery (OCD)}.}}
\label{Subsubsec:ocd}
{OCD extends conventional category discovery to an inductive learning paradigm with streaming inference. It trains on a labelled support set $D_S$ to cluster unlabelled query set $D_Q$ where $D_S$ is unavailable during training and its samples are individually at test time.}

SMILE~\cite{du2023on} first introduces On-the-Fly Category Discovery, leveraging hash-based descriptors and a sign-magnitude disentanglement architecture to mitigate intra-category variance.
Similarly, PHE~\cite{zheng2024prototypical} proposes a robust framework for fine-grained on-the-fly category discovery using hash-based representations with prototypical encoding, combining Category-aware Prototype Generation and Discriminative Hash Encoding.
DiffGRE~\cite{liu2025generate} adopts a Generate–Refine–Encode pipeline: cross-space slerp (Stable Diffusion + CLIP) synthesizes novel samples, a diversity-driven filter removes near-known artifacts, and semi-supervised leader prototypes enable high-dimensional online clustering.

%% file: tabs/OCD_table.tex
\begin{table*}
\centering
\caption{Summary of essential components for reviewed OCD approaches.}
\vspace{-5pt}
\label{table:OCD}
\begin{threeparttable}
	\resizebox{0.5\textwidth}{!}{
		\setlength\tabcolsep{6pt}
		\renewcommand\arraystretch{1.0}
		\begin{tabular}{|c|r||c|c|c|}
			\hline
			\rowcolor[HTML]{D3D3D3}
			Year & Method~~~~ & Pub. & Backbone & Label Assignment \\
			\hline
			\hline
            \hline
			\multirow{1}{*}{\rotatebox{0}{2023}}
            & SMILE~\cite{du2023on} & \textit{CVPR} & ViT & Hash-based \\
            \hline
			\multirow{1}{*}{\rotatebox{0}{2024}}
            & PHE~\cite{zheng2024prototypical} & \textit{NeurIPS} & ViT & Hamming Ball-Based \\
            \multirow{1}{*}{\rotatebox{0}{2024}}
            & DiffGRE~\cite{liu2025generate} & \textit{ICCV} & ViT & Online Clustering \\
            \hline
		\end{tabular}
        }
\end{threeparttable}
\vspace{-15pt}
\end{table*}

%% file: tabs/CDwDomain_shift_table.tex
\begin{table*}
\centering
\caption{
Summary of essential components for reviewed CD with Domain shift approaches.
}
\vspace{-6pt}
\label{table:CDwDS}
\begin{threeparttable}
	\resizebox{0.75\textwidth}{!}{
		\setlength\tabcolsep{6pt}
		\renewcommand\arraystretch{1.0}
		\begin{tabular}{|c|r||c|c|c|c|c|}
			\hline
			\rowcolor[HTML]{D3D3D3}
			Year & Method~~~~ & Pub. & Backbone & $\Omega_{\mathcal{U}}$ & Label Assignment & $\mathcal{Y_L} \cap \mathcal{Y_U}$\\
			\hline
			\hline
			\hline
			\multirow{2}{*}{{2022}}
                & Yu \etal~\cite{Yu_Ikami_Irie_Aizawa_2022} & \textit{AAAI} & ResNet & Single New Domain & Parametric Classifier & $\varnothing$\\
                & SCDA~\cite{zhuang2022opensetdomainadaptation} & \textit{ICME} & ResNet & Multiple New Domains & Parametric Classifier & $\varnothing$\\
            \hline
			\multirow{1}{*}{{2023}}
                & SAN~\cite{zang2023boostingnovelcategorydiscovery} & \textit{ICCV} & ResNet & Single New Domain & Parametric Classifier & $\varnothing$\\
            \hline
			\multirow{1}{*}{{2024}}
                & CDAD-Net~\cite{rongali2024cdadnetbridgingdomaingaps} & \textit{CVPRW} & ViT & Single New Domain & $k$-means & $ \neq \varnothing$\\
            \hline
			\multirow{1}{*}{{2025}}
                & HiLo~\cite{wang2024hilolearningframeworkgeneralized} & \textit{ICLR} & ViT & Multiple New Domains & Parametric Classifier & $ \neq \varnothing$\\
            \hline
			\multirow{1}{*}{{Preprint}}
                & Wang \etal~\cite{wang2024exclusivestyleremovalcross} & \textit{ArXiv} & ViT & Single New Domain & Parametric Classifier & $\varnothing$\\
            \hline
		\end{tabular}
        }
\end{threeparttable}
\vspace{-15pt}
\end{table*}

%% file: Parts/CDwdomainShift.tex
\noindent \underline{\textit{\textbf{Category Discovery with Domain Shift}.}}
\label{Subsecsec:cdwds}
Category Discovery with Domain Shift addresses the challenge of discovering novel categories when there is a discrepancy between the labelled source domain and the unlabelled target domain.
{Formally, the labelled data $\mathcal{D}_L$ is assumed to be exclusively drawn from the base domain $\Omega_{B}$, and {let $\mathcal{D}_U$ include samples from both $\Omega_{B}$ and a novel domain $\Omega_{N}$}. 
The objective is to accurately classify images from the combined domain $\Omega = \Omega_{B} \cup \Omega_{N}$, assuming that the base and novel domains are disjoint (\ie, $\Omega_{B} \cap \Omega_{N} = \varnothing$). In practice, the novel domain $\Omega_{N}$ may encompass multiple subdomains.}

Yu~\etal~\cite{Yu_Ikami_Irie_Aizawa_2022} propose a self-labelling framework for unsupervised domain adaptation, assuming unlabelled data contains only novel classes from a single target domain. It aligns source and target features via prototypical contrastive learning and mutual information maximization.
Zhuang~\etal~\cite{zhuang2022opensetdomainadaptation} propose the Self-supervised Class-Discovering Adapter (SCDA) for open set domain adaptation. SCDA pre-trains a model with adversarial learning to distinguish known from unknown target domain samples, and then alternates clustering-driven class discovery with self-supervised adaptation to dynamically update its classifier.
SAN~\cite{zang2023boostingnovelcategorydiscovery} uses soft contrastive learning to reduce domain-induced view noise and introduces an all-in-one classifier that handles both known and unknown categories. However, SAN assumes all unlabelled samples come from a single novel domain and only requires a general ``unknown'' label for novel categories, limiting its applicability to more complex settings.
Similarly, CDAD-Net~\cite{rongali2024cdadnetbridgingdomaingaps} adopts an entropy-driven adversarial learning approach to align representations between source and target domains, and refines representations using a conditional image inpainting. CDAD-Net also employs the elbow method~\cite{elbow} for class number estimation.
Wang~\etal~\cite{wang2024exclusivestyleremovalcross} propose a style removal module that disentangles domain-specific style information from domain-invariant content features, improving feature alignment.
HiLo~\cite{wang2024hilolearningframeworkgeneralized} handles multi-domain unlabelled data by disentangling semantic and domain-specific features, using PatchMix~\cite{hong2024patchmix} for embedding-level blending and curriculum learning~\cite{bengio2009curriculum} to gradually address complex domain shifts.
%
Among these methods, \cite{Yu_Ikami_Irie_Aizawa_2022, zhuang2022opensetdomainadaptation, zang2023boostingnovelcategorydiscovery, wang2024exclusivestyleremovalcross} are built upon the NCD setting assuming disjoint label spaces between $\mathcal{Y}_L$ and $\mathcal{Y}_U$, while others~\cite{wang2024hilolearningframeworkgeneralized, rongali2024cdadnetbridgingdomaingaps} are developed under the more general GCD setting.

%% file: Parts/LongTailCD.tex
\input{tabs/CDwLongTail_table}

\input{tabs/SCD_table}

\input{tabs/FCD_table}

\noindent \underline{\textit{\textbf{Distribution-Agnostic Category Discovery (DA-CD)}.}}
\label{Subsecsec:dacd}
%
DA-CD tackles the challenge of imbalanced class distributions by developing robust methodologies that do not assume uniformity in labelled or unlabelled data, thereby enabling the effective discovery of novel categories in realistic environments. 
{It acknowledges that the data may follow a skewed distribution, such that for certain categories \(\mathcal{Y}_i\) and \(\mathcal{Y}_j\) within the set \(\mathcal{Y}\), it holds that
$\mathbb{P}_{\mathcal{Y_x}}(\mathcal{Y}_i) > \mathbb{P}_{\mathcal{Y_x}}(\mathcal{Y}_j).$
In this formulation, the set \(\mathcal{Y_x}\) may refer to either the labelled categories \(\mathcal{Y}_L\) or the unlabelled categories \(\mathcal{Y}_U\).}

NCDLR~\cite{zhang2023novel} targets long-tailed imbalances with relaxed optimal-transport self-labeling and equiangular prototypes. A $k$-means-based approach~\cite{vaze2022gcd} is used to estimate the number of novel classes.
Similarly, BaCon~\cite{bai2023towards} introduces a dual-branch framework comprising a contrastive-learning branch that dynamically estimates the data distribution through clustering to enhance feature representations, alongside a pseudo-labelling branch that produces class predictions for both known and novel categories. 
To address distribution shifts and class imbalance in biomedical imaging, Fan~\etal~\cite{Fan_2024_CVPR} use von Mises–Fisher distributions on a hypersphere to decouple semantics from noise, with geometric constraints for better class separation. They estimate class numbers via spectral analysis of the graph Laplacian.
%
BYOP~\cite{yang2023bootstrap} extends the base NCD setting to handle novel classes with imbalanced distributions by iteratively refining the class prior using model predictions and clustering data based on the estimated distribution, with dynamically adjusted temperature settings to improve prediction confidence in uncertain categories.
ImbaGCD~\cite{li2023imbagcdimbalancedgeneralizedcategory} introduces an optimal transport–based expectation-maximization framework to iteratively align class priors, reducing head-class bias and improving representation and label assignment under long-tailed distributions.
GCDLR~\cite{li2024generalizedcategoriesdiscoverylongtailed} extends ImbaGCD by reweighting rare classes and enforcing class-prior constraints for better distribution alignment, but still assumes a known number of novel categories.
Zhao \etal~\cite{zhao2025generalized} propose a dual reliable-sample miner, plus a density-peak–driven estimator for the number of novel classes.

%


%% file: tabs/CDwLongTail_table.tex
\begin{table*}
\centering
\caption{Summary of essential components for reviewed DA-CD approaches.
}
\vspace{-6pt}
\label{table:DACD}
\begin{threeparttable}
	\resizebox{0.72\textwidth}{!}{
		\setlength\tabcolsep{6pt}
		\renewcommand\arraystretch{1.0}
		\begin{tabular}{|c|r||c|c|c|c|c|}
			\hline
			\rowcolor[HTML]{D3D3D3}
			
			Year &Method~~~~  &Pub. &Backbone  
			& Scenario & Label Assignment 
			  & $\mathcal{Y_L} \cap \mathcal{Y_U}$\\
			\hline
			\hline
			\hline
			\multirow{5}{*}{{2023}}
                &NCDLR~\cite{zhang2023novel}&\textit{TMLR}&ViT&Long-tailed Distribution for $\mathcal{Y_L} \&\mathcal{Y_U}$&Parametric Classifier&$\varnothing$\\
                &ImbaGCD~\cite{li2023imbagcdimbalancedgeneralizedcategory}&\textit{CVPRW}&ResNet&Imbalanced Distribution for $\mathcal{Y_U}$&Parametric Classifier&$\neq \varnothing$\\
                &GCDLR~\cite{li2024generalizedcategoriesdiscoverylongtailed}&\textit{ICCVW}&ResNet&Imbalanced Distribution for $\mathcal{Y_U}$&Parametric Classifier&$\neq \varnothing$\\
                &BYOP~\cite{yang2023bootstrap}&\textit{CVPR}&ResNet&Imbalanced Distribution for $\mathcal{Y_U}$&Parametric Classifier&$\varnothing$\\
                &BaCon~\cite{bai2023towards}&\textit{NeurIPS}&ViT&Long-tailed Distribution for $\mathcal{Y_L} \&\mathcal{Y_U}$&$k$-means&$\neq \varnothing$\\
                \hline
			\multirow{1}{*}{{2024}}
                &Fan \etal~\cite{Fan_2024_CVPR}&\textit{CVPR}&ViT&Long-tailed Distribution for $\mathcal{Y_L} \&\mathcal{Y_U}$&Spectral graph&$\neq \varnothing$\\
                \hline
			\multirow{1}{*}{{Preprint}}
                &Zhao \etal~\cite{zhao2025generalized}&\textit{ArXiv}&ViT&Long-tailed Distribution for $\mathcal{Y_L} \&\mathcal{Y_U}$&Density-based Clustering&$\neq \varnothing$\\
                \hline

		\end{tabular}
        }
\end{threeparttable}
\vspace{-15pt}
\end{table*}

%% file: tabs/SCD_table.tex
\begin{table*}[h!]
\centering
\caption{Summary of essential components for reviewed SCD approaches.}
\vspace{-6pt}
\label{table:SCD}
\begin{threeparttable}
	\resizebox{0.77\linewidth}{!}{
		\setlength\tabcolsep{6pt}
		\renewcommand\arraystretch{1.0}
		\begin{tabular}{|c|r||c|c|c|c|c|c|}
			\hline
			\rowcolor[HTML]{D3D3D3}
			
			Year &Method~~~~  &Pub. &Backbone  
			& Word Space & Label Assignment 
			& \# Unlabelled categories & $\mathcal{Y_L} \cap \mathcal{Y_U}$ \\
			\hline
			\hline
                \hline
                \multirow{3}{*}{{2024}}
                &SCD~\cite{han2024whatsnameclassindices}&\textit{CVPRW}&ViT&$\sim$Open&$k$-means+Top-$k$ Voting& Known & $\neq  \varnothing$\\
                &SNCD~\cite{Wang_Lei_Chen_Liu_2024}&\textit{AAAI}&ResNet&$\mathcal{C}_{base} + \mathcal{C}_{novel}$&Parametric Classifier&Known&$\varnothing$\\
                &FineR~\cite{liu2024democratizing}&\textit{ICLR}&ViT&$\sim$Open&Parametric Classifier&Unknown&$\neq  \varnothing$\\

            \hline
		\end{tabular}
        }
\end{threeparttable}
\vspace{-10pt}
\end{table*}

%% file: tabs/FCD_table.tex
\begin{table*}
\centering
\caption{Summary of essential components for reviewed FCD approaches.
}
\vspace{-6pt}
\label{table:FCD}
\begin{threeparttable}
	\resizebox{0.55\textwidth}{!}{
		\setlength\tabcolsep{6pt}
		\renewcommand\arraystretch{1.0}
		\begin{tabular}{|c|r||c|c|c|c|c|}
			\hline
			\rowcolor[HTML]{D3D3D3}
			
			Year &Method~~~~  &Pub. &Backbone  
			 & Label Assignment 
			 & $\mathcal{Y_L} \cap \mathcal{Y_U}$\\
			\hline
			\hline
            \hline
			\multirow{1}{*}{{2023}}
            &FedoSSL~\cite{zhang2023unbiasedtrainingfederatedopenworld}&\textit{ICML}&ResNet&Parametric Classifier&$\neq  \varnothing$\\
			\hline
			\multirow{1}{*}{{2024}}
            &FedGCD~\cite{pu2023federatedgeneralizedcategorydiscovery}&\textit{CVPR}&ViT&GMM&$\neq  \varnothing$\\
			\hline
			\multirow{1}{*}{{2025}}
            &GAL~\cite{wang2023federatedcontinualnovelclass}&\textit{ICCV}&ResNet&Parametric Classifier&$\varnothing$\\
            \hline
		\end{tabular}
        }
\end{threeparttable}
\vspace{-10pt}
\end{table*}

%% file: Parts/SCD.tex

\noindent
\underline{\textit{\textbf{Semantic Category Discovery (SCD)}.}}
\label{Subsecsec:scd}
Unlike conventional category discovery methods, which only group visually similar samples, SCD extends GCD by directly discovering semantic class names through open-vocabulary reasoning.
%
{Specifically, SCD leverages an open vocabulary space to achieve this goal. {In this context—such as an open-vocabulary setting—WordNet~\cite{wordnet}, which comprises approximately $68,000$ labels, is used as a comprehensive and unconstrained lexical resource to facilitate the assignment of semantic labels.}}
%

Wang~\etal~\cite{Wang_Lei_Chen_Liu_2024} introduce Semantic-guided Novel Category Discovery, which supplements NCD by mapping semantic labels to unlabelled data instead of using discrete one-hot labels. This is achieved by combining a semantic similarity matrix with dynamic visual prototypes, enabling the joint optimization of classification and clustering via mutual information maximization.
%
%
In a more challenging setting, Han~\etal~\cite{han2024whatsnameclassindices} introduce an SCD framework where the label space comprises an open vocabulary. 
By leveraging VLMs such as CLIP~\cite{clip}, their approach aligns visual and textual data to facilitate more effective label assignment from an expansive vocabulary. 
A voting mechanism within clusters then identifies the most probable semantic labels, and an iterative refinement process continuously enhances label accuracy as the model evolves.
FineR~\cite{liu2024democratizing} proposes a training-free framework for fine-grained visual recognition, aligning with the definition used in SCD. It utilises a VQA model (\eg, BLIP-2~\cite{li2023blip}) to extract visual attributes, an LLM (\eg, GPT~\cite{brown2020language}) to generate candidate labels, and a VLM to build a multimodal classifier for zero-shot recognition.


%% file: Parts/few-shotsCD.tex
\noindent \underline{\textit{\textbf{Few-Shot Category Discovery (FS-CD)}.}}
\label{Subsecsec:fs-cd}
FS-CD tackles the challenge of identifying novel categories with limited labelled data by leveraging prior knowledge from known classes and maximizing the value of scarce annotations. 
{It extends conventional category discovery by incorporating few-shot learning principles. In particular, FS-CD adopts an $N$-way, $k$-shot framework requiring category discovery with only $k$ labelled examples per base class.}

Chi~\etal~\cite{chi2022meta} extend NCD to a few-shot setting based on the assumption that both base and novel categories share high-level semantic features. 
Their approach adapts meta-learning techniques - Model-Agnostic Meta-Learning (MAML)~\cite{finn2017model} and Prototypical Networks (ProtoNet)~\cite{snell2017prototypical}, shifting the focus from classification to clustering tasks—crucial for few-shot category discovery. 
A key innovation of their method is the clustering-rule-aware task sampler, which ensures consistent clustering rules during training, thereby enhancing the model's ability to generalize to novel categories despite limited labelled data. However, this method assumes prior knowledge of the number of novel categories.

%% file: Parts/FCD.tex
\noindent \underline{\textit{\textbf{Federated Category Discovery (FCD)}.}}
\label{Subsecsec:fcd}
%
FCD addresses the challenge of collaboratively discovering novel categories from decentralized, privacy-preserving datasets with heterogeneous label distributions, requiring methods that integrate diverse local knowledge while preserving data privacy. {It considers $N^L$ local clients, each equipped with its own private labelled dataset $\mathcal{D}_n^L = \{(\mathbf{x}_i, y_i)\} \subset \mathcal{X}_n^L \times \mathcal{Y}_n^L$ and an unlabelled dataset $\mathcal{D}_n^U = \{(\mathbf{x}_i, \hat{y}_i)\}_{i=1}^K \subset \mathcal{X}_n^U \times \mathcal{Y}_n^U$, where the local label space is defined as $\mathcal{Y}_n = \mathcal{Y}_n^L \cup \mathcal{Y}_n^U$. Notably, the label spaces across different clients are not assumed to be identical (\ie, for any two clients $i$ and $j$, $\mathcal{Y}_i$ may differ from $\mathcal{Y}_j$), thereby reflecting heterogeneous data distributions. The primary objective of FCD is to collaboratively train a generalized global model that aggregates knowledge from all local clients, without exposing any private data.}

FedoSSL~\cite{zhang2023unbiasedtrainingfederatedopenworld} extends GCD to decentralized learning. It introduces an uncertainty-aware regulation loss to balance learning rates between locally and globally unseen categories, and employs a calibration module for consistent model aggregation. However, FedoSSL assumes prior knowledge of the total number of categories across clients, limiting its use in settings that require category estimation.
Fed-GCD~\cite{pu2023federatedgeneralizedcategorydiscovery} introduces associated Gaussian contrastive learning using a learnable GMM. A client semantics association mechanism aggregates local GMMs into a global GMM on the central server, facilitating shared semantic knowledge discovery. It also integrates semi-supervised FINCH~\cite{sarfraz2019finch} to estimate the number of novel categories per client.
Wang~\etal~\cite{wang2023federatedcontinualnovelclass} present the GAL framework for FCD in a continual setting. GAL combines modified prototype contrastive learning for robust representation learning, prototype-based filtering for accurate label assignment, and the \textit{potential prototype merge} algorithm to estimate the number of novel categories.



%% file: Parts/OthersCD.tex
\input{tabs/othersCD_table}
\noindent
\underline{\textit{\textbf{Category Discovery in Scene Understandings}.}}
%
Zhao~\etal~\cite{zhao2022ncdss} extend category discovery to semantic segmentation. They employ a pretrained BASNet~\cite{qin2019basnet} saliency model alongside a base segmentation model trained on labelled base categories to isolate novel regions. Pseudo-labels are assigned via $k$-means clustering. 
To mitigate clustering noise, they introduce entropy-based uncertainty modeling and self-training, refining pseudo-labels through entropy ranking and dynamic data reassignment.

%

Zhong~\etal~\cite{osodd2022} propose Open-Set Object Detection and Discovery (OSODD), which jointly detects known objects and discovers novel categories. Their two-stage framework first employs an open-set detector to identify both known and unknown objects, followed by contrastive learning and semi-supervised clustering to group unknown objects into novel classes.
DRM~\cite{feng2024debiasednovelcategorydiscovering} tackles the bias toward known categories in NCD for object detection. It employs a dual Region Proposal Network (RPN) strategy that combines class-aware and class-agnostic RPNs, enhancing feature representations with semi-supervised instance-level contrastive learning. 
%
PANDAS~\cite{hayes2024pandasprototypebasednovelclass} introduces a prototype-based approach for novel class discovery and detection on top of Faster R-CNN. At inference, objects are classified by comparing feature vectors to both base and novel prototypes via a similarity metric, enabling detection without prior knowledge of the number of novel classes.
Hoang~\etal~\cite{hoang2025generalized} propose a GCD method for instance segmentation that tackles novel category discovery and class imbalance. The approach employs instance-wise temperature assignment for contrastive learning, soft attention modules to enhance object-specific features, and reliability-based dynamic learning to iteratively refine pseudo-labels with class-wise thresholds.
%

\noindent
\underline{\textit{\textbf{Category Discovery for Videos}.}}
%
NEV~\cite{hasan2023nevncdnegativelearningentropy} tackles NCD in video action recognition by employing negative learning, entropy regularization, and variance regularization. Additionally, it incorporates view-invariant learning through adversarial training to filter viewpoint-specific information and contrastive learning to align representations of the same action from different viewpoints.
%

\noindent
\underline{\textit{\textbf{Category Discovery for 3D Objects}.}}
NOPS~\cite{riz2023novel} extends category discovery to 3D point cloud under the NCD setting. It leverages online clustering and uncertainty quantification to generate pseudo-labels for novel categories, and further improves feature space partitioning through a class-balanced queuing strategy and a multi-headed segmentation technique.
%
Similarly, Xu~\etal~\cite{xu2024dual} propose a dual-level adaptive self-labelling framework for 3D point cloud segmentation. It addresses class imbalance and spatial context by leveraging point-level and region-level representations. A key contribution is a dynamic semi-relaxed optimal transport algorithm that adjusts regularization to handle imbalanced novel classes.
%
DNIK~\cite{NEURIPS2023_aa31eee8} presents a different approach by utilizing geometric part-based features from known classes. It constructs a part concept bank to represent 3D shapes and uses a \textit{part relation encoder} to capture spatial relationships.
%
Notably, NOPS~\cite{riz2023novel} and DNIK~\cite{NEURIPS2023_aa31eee8} assume the number of novel categories is known in advance, whereas Xu~\etal~\cite{xu2024dual} employ an off-the-shelf $k$-means-based approach for category estimation.

\noindent
\underline{\textit{\textbf{Category Discovery for Text}.}}
DPN~\cite{an2023generalized} introduces a \textit{decoupled prototypical network} that separates the learning of known and novel categories. It employs bipartite matching for prototype alignment between labelled and unlabelled data, and \textit{semantic-aware prototypical learning} to enhance feature discrimination and reduce pseudo-label noise.
An~\etal~\cite{an-etal-2022-fine} propose Fine-grained Category Discovery under Coarse-grained supervision (FCDC), using a hierarchical model based on pretrained BERT~\cite{devlin2019bertpretrainingdeepbidirectional} to learn coarse and fine-grained features, aided by a self-contrastive module to balance class distances.
%
TAN~\cite{an2024transfer} presents a prototype-based framework in the GCD setting, incorporating knowledge transfer mechanisms and feature alignment strategies to address noisy unlabelled data. It utilises a pretrained BERT~\cite{devlin2019bertpretrainingdeepbidirectional} model and a filtering algorithm for novel category estimation.
%
KTN~\cite{Shi_An_Tian_Chen_Wu_Wang_Chen_2024} applies an entropy-based technique to identify novel category samples, combining prototype-based semantic similarities and logit adjustments for improved pseudo-label accuracy.
%
SDC~\cite{an2024unleashing} enhances knowledge transfer using a dual logit adjustment strategy to reduce bias and facilitate transfer between similar known and novel categories.
%
KNoRD~\cite{hogan-etal-2023-open} extends category discovery to open relation extraction in long-tailed NCD, combining prompt-based training and GMM clustering for semantically aligned relation representations.
%
SFGRD~\cite{wang2024learningsemifactualsdebiasedsemanticaware} proposes a two-stage framework for open-world relation extraction, reducing entity and context biases while enhancing relational semantics through a cluster-semantic space and iterative novel relation estimation via clustering.

\noindent
\underline{\textit{\textbf{Category Discovery for Tabular Data}.}}
TabularNCD~\cite{tr2022method} extends category discovery to tabular data under the NCD setting, addressing challenges from its heterogeneous structure. It employs self-supervised learning to pre-train latent representations on both labelled and unlabelled data, jointly optimizing a classification task on known classes and a clustering task to discover novel categories.
%
Colin~\etal~\cite{tr2023interface} develop an interactive interface for category discovery in tabular data, enabling subject-matter experts to analyse results without machine learning expertise. The system integrates NCD methods, such as TabularNCD~\cite{tr2022method}, with traditional clustering algorithms involving $k$-means and spectral clustering.
%
PBT~\cite{tr2023practical} enhances unsupervised clustering methods such as $k$-means++ and Spectral Clustering—by incorporating labelled data from known classes to guide the clustering process. 
It also leverages \textit{cluster validity indices} computed in the latent feature space to estimate the number of novel classes.

\noindent
\underline{\textit{\textbf{Category Discovery for Graph}.}}
SWORD~\cite{hou2024ncncdnovelclassdiscovery} introduces category discovery for graph classification. It employs self-training to learn from unlabelled nodes, uses prototype replay to preserve knowledge of existing classes, and incorporates knowledge distillation to mitigate catastrophic forgetting.

\noindent
\underline{\textit{\textbf{Category Discovery Cross Modalities}.}}
Jia~\etal~\cite{Jia2021JointRL} introduce a unified framework for multimodal NCD by combining instance and category contrastive learning with cross-modal contrastive learning. 
%
Zhou~\etal~\cite{zhou2024novel} extend UNO~\cite{uno} to a multimodal framework for chest X-rays. They first apply multimodal contrastive learning to fine-tune image and text encoders, then fuse pseudo-labels from both modalities to improve novel class discovery.
%
Furthermore, Su~\etal~\cite{su2024multimodalgeneralizedcategorydiscovery} extend GCD to multimodal datasets by employing contrastive learning to align feature spaces across modalities and cross-modal distillation to ensure consistent classification, while incorporating a prototypical classifier for effective category partitioning.

%% file: tabs/othersCD_table.tex
\begin{table*}
\centering
\caption{Summary of essential components for reviewed CD beyond images.}
\vspace{-6pt}
\label{table:othersCD}
\begin{threeparttable}
	\resizebox{0.74\linewidth}{!}{
		\setlength\tabcolsep{6pt}
		\renewcommand\arraystretch{1.0}
		\begin{tabular}{|c|c|r||c|c|c|}
			\hline
			\rowcolor[HTML]{D3D3D3}
			Task & Year & Method~~~~ & Pub. & Backbone & Label Assignment \\
			\hline
			\hline
            \hline
            \multirow{5}{*}{\rotatebox{0}{\textit{Seg.\&Dect.}}}
            &\multirow{2}{*}{2022}
            & Zhao \etal~\cite{zhao2022ncdss} & \textit{CVPR} & ResNet & $k$-means \\
            && Zhong \etal~\cite{osodd2022} & \textit{CVPRW} & ResNet & $k$-means \\
            \cline{2-6}
            &\multirow{2}{*}{2024}
            & DRM~\cite{feng2024debiasednovelcategorydiscovering} & \textit{AAAI} & Faster R-CNN & $k$-means \\
            && PANDAS~\cite{hayes2024pandasprototypebasednovelclass} & \textit{CoLLAs} & Faster R-CNN & $k$-means \\
            \cline{2-6}
            & 2025 & Hoang \etal~\cite{hoang2025generalized} & \textit{AAAI} & ResNet+SAM & Parametric Classifier \\
            \hline
            \multirow{1}{*}{\rotatebox{0}{\textit{Videos}}}
            & 2023 & NEV~\cite{hasan2023nevncdnegativelearningentropy} & \textit{ICIP} & SlowFast, R2+1D, P3D & Parametric Classifier \\
            \hline
            \multirow{3}{*}{\rotatebox{0}{\textit{3D.}}}
            &\multirow{2}{*}{2023}
            & NOPS~\cite{riz2023novel} & \textit{CVPR} & MinkowskiUNet & $k$-means \\
            && DNIK~\cite{NEURIPS2023_aa31eee8} & \textit{NeurIPS} & PointNet & Parametric Classifier \\
            \cline{2-6}
            & 2024 & Xu \etal~\cite{xu2024dual} & \textit{ECCV} & MinkowskiUNet & Parametric Classifier \\
            \hline
            \multirow{7}{*}{\rotatebox{0}{\textit{Text}}}
            & 2022 & An \etal~\cite{an-etal-2022-fine} & \textit{EMNLP} & BERT & $k$-means \\
            \cline{2-6}
            &\multirow{2}{*}{2023}
            & DPN~\cite{an2023generalized} & \textit{AAAI} & BERT & $k$-means \\
            && KNoRD~\cite{hogan-etal-2023-open} & \textit{EMNLP} & DeBERTa & GMM \\
            \cline{2-6}
            &\multirow{2}{*}{2024}
            & TAN~\cite{an2024transfer} & \textit{AAAI} & BERT & $k$-means \\
            && KTN~\cite{Shi_An_Tian_Chen_Wu_Wang_Chen_2024} & \textit{AAAI} & BERT & GMM \\
            \cline{2-6}
            & 2025 & SDC~\cite{an2024unleashing} & \textit{AAAI} & BERT & Parametric Classifier \\
            \cline{2-6}
            & preprint & SFGRD~\cite{wang2024learningsemifactualsdebiasedsemanticaware} & \textit{ArXiv} & BERT & $k$-means \\
            \hline
            \multirow{3}{*}{\rotatebox{0}{\textit{Tabular}}}
            & 2022 & TabularNCD~\cite{tr2022method} & \textit{ICKG} & \makecell{Encoder\\(2 dense hidden layers)} & Parametric Classifier \\
            \cline{2-6}
            &\multirow{2}{*}{2023}
            & PBT~\cite{tr2023practical} & \textit{DMKN} & PBN & $k$-means \\
            && Colin \etal~\cite{tr2023interface} & \textit{ECML PKDD} & N/A & $k$-means \\
            \hline
            \multirow{1}{*}{\rotatebox{0}{\textit{Graph}}}
            & 2024 & SWORD~\cite{hou2024ncncdnovelclassdiscovery} & \textit{CIKM} & GCN, GAT, GraphSAGE & Parametric Classifier \\
            \hline
            \multirow{2}{*}{\rotatebox{0}{\textit{Multi-Modal}}}
            & 2021 & Jia \etal~\cite{Jia2021JointRL} & \textit{ICCV} & ResNet, R3D-18 & Parametric Classifier \\
            \cline{2-6}
            & 2024 & Zhou \etal~\cite{zhou2024novel} & \textit{AAAI} & ResNet\&BioClinicalBERT & Parametric Classifier \\
            \cline{2-6}
            & preprint & Su \etal~\cite{su2024multimodalgeneralizedcategorydiscovery} & \textit{AAAI} & ViT & Parametric Classifier \\
            \hline
		\end{tabular}
        }
\end{threeparttable}
\vspace{-15pt}
\end{table*}

%% file: Secs/4_benchmark.tex
\input{tabs/datasets_split}

\input{tabs/benchmark_ncd}
\section{Benchmark}
\label{Sec:benchmark}
In this section, we review commonly used datasets and their construction methods, outline the evaluation protocols, and consolidate the performance of the reviewed category-discovery methods on the two base settings-\textit{NCD} and \textit{GCD}. For each task, we select the most commonly used datasets for benchmarking performance. Performance scores are sourced from the original articles unless otherwise specified. More in-depth descriptions of the other datasets are included in the Appendix-V.

\vspace{-8pt}
\subsection{Dataset}
\label{Subsec:dataset}
\vspace{-2pt}
In category discovery, a diverse range of datasets is used, spanning \textit{generic} and \textit{fine-grained} categories across various domains, including natural images, plant species, and biomedical fields.
In this subsection, we introduce six widely utilised datasets, comprising three generic and three fine-grained datasets, along with the dataset-splitting methods commonly applied in the field.
{The statistics for the commonly used datasets are summarized in Tab~\ref{tab:datasplit}.}

The most widely used \textit{generic} datasets:
\textit{\textbf{ImageNet-1K\&100}}~\cite{russakovsky2015imagenet} is a widely used dataset for natural image classification. ImageNet-$1$K contains over one million labelled images across $1,000$ object categories. ImageNet-$100$ is a subset with $100$ randomly selected classes.
\textit{\textbf{CIFAR-10 \& CIFAR-100}}~\cite{krizhevsky2009cifar} are both natural images sized in $32 \times 32$. CIFAR-$10$ contains $50,000$ images spanning across $10$ different classes and CIFAR-$100$ includes $100$ classes, with each class containing $500$ images. 


{The most widely used \textit{fine-grained} benchmark is \textbf{SSB}~\cite{vaze2022opensetSSB}, which includes three datasets: CUB~\cite{WahCUB_200_2011}, Stanford-Cars~\cite{stanfordCars}, and FGVC-Aircraft~\cite{aircrafts}.}
\textit{\textbf{CUB}}~\cite{WahCUB_200_2011} is a widely used fine-grained dataset, particularly focused on bird species recognition.
\textit{\textbf{Stanford-Cars}}~\cite{stanfordCars} is a large-scale dataset designed for fine-grained vehicle classification tasks.
\textit{\textbf{FGVC-Aircraft}}~\cite{aircrafts} is a fine-grained visual classification dataset focused on aircraft recognition. 

Other Datasets:
\textit{\textbf{Caltech-101}}~\cite{fei2006one} is a collection of images across 101 object categories, and it is re-structured for CCD.
\textit{\textbf{TinyImageNet}}~\cite{le2015tinyImagenet} is a reduced version of ImageNet with $200$ classes, each having $500$ images, all resized to $64\times64$ pixels for computational efficiency.

\textit{Dataset Splitting}: In category discovery, categories are typically divided into base and novel categories. Given the total number of categories as $|\mathcal{C}|$, the number of base categories and novel categories are denoted as $|\mathcal{C_B}|$ and $|\mathcal{C_N}|$ respectively with the constraint $|\mathcal{C_B}| + |\mathcal{C_N}| = |\mathcal{C}|$. 
The common splitting ratio of base and novel categories is $|\mathcal{C_B}|:|\mathcal{C_N}| = 5:5$, except for CIFAR-100, which typically uses a $|\mathcal{C_B}|:|\mathcal{C_N}| = 8:2$ ratio. 
In GCD, since the unlabelled data contains both base and novel categories, it is necessary to further partition the base categories into labelled and unlabelled subsets. 
The typical ratio for this split is $5:5$, although a $1:9$ ratio is sometimes used for more challenging cases~\cite{rizve2022towards, orca2022}.
In Tab.~\ref{tab:datasplit}, we present detailed statistics of commonly used datasets for GCD. 



\vspace{-10pt}
\subsection{Evaluation Protocol}
\label{SubSec:eva_proto}
To evaluate the performance of category discovery, the predominant metric employed is the \textit{Clustering Accuracy} on unlabelled dataset $\mathcal{D}_U$. This metric is formulated as
\vspace{-5pt}
\begin{equation}
\vspace{-5pt}
\nonumber
    ACC = \max_{p \in \mathcal{P(Y_U)}}
    \frac{1}{M}
    \sum_{i=1}^{M}
    \mathds{1}\{y_i = p(\hat{y}_i)\},
\end{equation}
where $M = |\mathcal{D}_U|$ denotes the total number of samples in $\mathcal{D}_U$, and $\mathcal{P}(\mathcal{Y}_U)$ represents the set of all possible permutations of the class labels. In practice, the Hungarian algorithm~\cite{kuhn1955hungarian} is employed to assign the optimal label permutations.

Within the realm of category discovery, three evaluation protocols are adopted to compute clustering accuracy:
\begin{enumerate}
    \item \textit{Task-aware Protocol:} This protocol applies the Hungarian algorithm separately to the novel category set $\mathcal{C}_N$ and the base category set $\mathcal{C}_B$. Originally derived from NCD settings—where the unlabelled data consist solely of novel categories (\ie, $\mathcal{C}_N = \mathcal{Y}_U$)—this approach has been extended in GCD~\cite{vaze2022gcd} to compute optimal label assignments independently for both sets.
    \item \textit{Task-agnostic Protocol:} This protocol computes the optimal label assignment over the entire label space $\mathcal{Y}_U$, which includes both novel and base categories, which is the most commonly adopted protocol for GCD. 
    
\end{enumerate}

It is important to note that the majority of category discovery methods operate in a transductive setting, where the evaluation is conducted on the same unlabelled dataset used in training. Nevertheless, some approaches~\cite{ma2024active} also report performance under an inductive setting using an unseen test set.

\input{tabs/benchmark_gcd}

\noindent
\textbf{\textit{Specific Evaluation Protocol for CCD}.}
In the continual learning setting, GM~\cite{zhang2022growmergeunifiedframework} introduces two evaluation metrics, $\mathcal{M}_f$ and $\mathcal{M}_d$, to measure the degree of forgetting in base categories and the ability to recognize novel categories. Specifically, $\mathcal{M}_f = \max_{t \in {1, \cdots, T}} \left| ACC^0_{\mathcal{B}} - ACC^t_{\mathcal{B}} \right|$ represents the maximum difference in clustering accuracy for base categories over time, while $\mathcal{M}_d = ACC^T_{\mathcal{N}}$ denotes the final clustering accuracy on novel categories, where $ACC$ is evaluated under the task-agnostic protocol. Additionally, PromptCCD~\cite{cendra2024promptccd} introduces \textit{continual ACC (cACC)}, which differs from previous metrics by considering not only the initial labelled data but also incorporating unlabelled data from previous timesteps along with their assigned labels. This allows \textit{cACC} to evaluate performance continuously across all timesteps, rather than focusing solely on the current timestep.

\vspace{-10pt}
\subsection{NCD Performance Benchmarking}
\label{Subsec:NCD_benchmark}

We select three fine-grained datasets: CUB~\cite{WahCUB_200_2011}, Stanford-Cars~\cite{stanfordCars}, and FGVC-Aircraft~\cite{aircrafts}. For these datasets, all results are generated using a DINO~\cite{caron2021dino} pre-trained ViT as the backbone. Regarding the generic datasets, we present results on CIFAR-10~\cite{krizhevsky2009cifar}, CIFAR100-20~\cite{krizhevsky2009cifar}, CIFAR100-50~\cite{krizhevsky2009cifar}, ImageNet-100~\cite{russakovsky2015imagenet}, and ImageNet-1K~\cite{russakovsky2015imagenet} for benchmarking the NCD algorithms. Most methods utilise ResNet-18~\cite{resnet} as the backbone for these datasets, with the exception of KCL~\cite{Hsu18_L2C}, DTC~\cite{Han2019DTC} and OpenMix~\cite{openmix2020}, which implement a six-layer VGG-like architecture~\cite{simonyan2014very} specifically on CIFAR100-20. 
Tab.~\ref{tab:ncd_results} displays the performance of 17 state-of-the-art NCD methods. The performance is assessed by the average $ACC$ across three runs on the `New' categories utilizing the task-aware evaluation protocol introduced in Sec.~\ref{SubSec:eva_proto}. For most results, we also report the variance of $ACC$ as shown in the included papers.




\vspace{-10pt}

\input{tabs/benchmark_gcd_dinov2}
\subsection{{GCD Performance Benchmarking}}
\label{Subsec:GCD_benchmark}
For the GCD task, we select the most widely-used fine-grained datasets, including CUB~\cite{WahCUB_200_2011}, Stanford-Cars~\cite{stanfordCars} and FGVC-Aircraft~\cite{aircrafts} as well as generic datasets including CIFAR-10~\cite{krizhevsky2009cifar}, CIFAR-100~\cite{krizhevsky2009cifar} and ImageNet-100~\cite{russakovsky2015imagenet} for benchmarking the GCD algorithms. 
Tab.~\ref{tab:gcd_results} shows the performance of $35$ state-of-the-art GCD methods on the six datasets based on DINO~\cite{caron2021dino} pre-trained ViT-B~\cite{dosovitskiy2020image} backbone. 
The performance is measured by $ACC$ across the `All', `Old' and `New' categories, as introduced in Sec.~\ref{SubSec:eva_proto}. Additionally, we report the results of selected GCD methods on the same datasets using ViT-B backbones pre-trained with DINOv2~\cite{oquabdinov2} and CLIP~\cite{clip} on Tab.~\ref{tab:gcd_results_dino2}.

Notably, several methods~\cite{orca2022, rizve2022openldn, rizve2022towards, ijcai2023p445, sun2024graph, wang2023discover, sun2023opencon, lps_ijcai2024, niu2024owmatch, Xiao_2024_CVPR} employ different data-splitting protocols or different backbones, resulting in inconsistent labelled and unlabelled sets and ultimately hindering fair comparisons. Therefore, we reproduce OwMatch~\cite{niu2024owmatch} as a representative method, with selection details in the Appendix.




\vspace{-8pt}
\subsection{{Benchmarking Takeaways}}


Based on the above benchmarking and analysis, we summarize the following takeaways:


\noindent 
\textbf{(i) Part-level information plays a critical role in category discovery, particularly in fine-grained settings where global features struggle to distinguish visually similar classes.}
{GCD~\cite{vaze2022gcd} shows that part-level cues naturally emerge in ViT attention heads and become more semantically specialized after contrastive fine-tuning, with certain heads attending to concrete parts such as {windshields, headlights}, or {wheelhouses} in a car.}
In light of this, GCD methods like~\cite{wang2024sptnet} and~\cite{dai2025adaptive} explicitly model localized object parts or discriminative regions, demonstrating stronger cues for separating subtle inter-class differences. Meanwhile, DualRS~\cite{dualRs} has also utilised the local parts for knowledge transfer in the NCD context. 
{As shown in Tab.~\ref{table:GCD}, SPTNet~\cite{wang2024sptnet} augments SimGCD~\cite{wen2023simgcd} with visual prompts to enhance patch-level discriminative cues, lifting \textit{All} accuracy from $60.3$ to $65.8$, $53.8$ to $59.0$, and $54.2$ to $59.3$ for fine-grained datasets respectively; building on this, Dai \etal~\cite{dai2025adaptive} employ adaptive part learning and further improve to $68.5$, $62.3$ and $60.9$.}

\noindent
\break \hfill 
\textbf{(ii) Large-scale self-supervised pretrained vision backbones provide a strong advantage for category discovery.}
%
%
{Early methods often train from scratch using self-supervised strategies such as RotNet~\cite{gidaris2018unsupervised} and MoCo~\cite{moco}.} DINO~\cite{caron2021dino} brings a clear improvement over previous methods, and DINOv2~\cite{oquabdinov2} further surpasses DINO by over $10\%$ on our benchmarks (Tab.~\ref{table:NCD} and Tab.~\ref{table:GCD}), underscoring the critical role of high-quality visual representations. 
These results highlight that stronger pretrained models not only improve feature quality but also enhance the robustness and scalability of GCD methods.
%
%
\noindent
\break \hfill
{
\textbf{(iii) Hierarchical information has been shown to be an important inductive bias for category discovery.}
SelEx~\cite{RastegarECCV2024} employs hierarchical classification, yielding a $17.9\%$ gain over the GCD baseline~\cite{vaze2022gcd}, {and SEAL~\cite{he2025seal} explicitly models hierarchies via coarse-to-fine labels, boosting the SimGCD baseline~\cite{wen2023simgcd} by $8.4\%$ on fine-grained datasets (see Tab.~\ref{tab:gcd_results}).}
%
%
HypCD~\cite{liu2025hyperbolic} encodes hierarchy implicitly through hyperbolic embeddings, and {further increases over SelEx by $6.5\%$} as shown on Tab.~\ref{tab:gcd_results}, underscoring the value of incorporating hierarchical information into the discovery process. 
}
\noindent
\break \hfill
{
\textbf{(vi) Auxiliary information beyond raw image features also greatly benefits category discovery.} 
Methods like MOS~\cite{peng2025mos}, GET~\cite{wang2024unlockingmultimodalpotentialclip}, and TextGCD~\cite{zheng2024textualknowledgematterscrossmodality} report measurable gains (see Tab.~\ref{tab:gcd_results_dino2}), and collectively demonstrate that incorporating additional signals, whether object-level or textual information, provides valuable semantic grounding that complements visual learning.
}

%% file: tabs/datasets_split.tex

\begin{table}[h]   
    \caption{Detailed Dataset splitting statistics for GCD. FG indicates whether the dataset is fine-grained.
    }
    \label{tab:datasplit}
    \centering
    \begin{tabular}{lc|rr|rr}
    \hline  
\rowcolor[HTML]{D3D3D3}
            &    & \multicolumn{2}{c|}{Labelled}  & \multicolumn{2}{c}{Unlabelled}\\
\rowcolor[HTML]{D3D3D3}    Dataset         & FG   & \#Image   & \#Class   & \#Image   & \#Class \\
    \hline \hline
    ImageNet-$100$~\cite{russakovsky2015imagenet}    & \textcolor{customred}{\ding{55}}   & $31.9$K     & $50$        & $95.3$K     & $100$ \\
    CIFAR-$10$~\cite{krizhevsky2009cifar}         & \textcolor{customred}{\ding{55}}    & $12.5$K     & $5$         & $37.5$K     & $10$ \\
    CIFAR-$100$~\cite{krizhevsky2009cifar}        & \textcolor{customred}{\ding{55}}    & $20.0$K     & $80$        & $30.0$K     & $100$ \\
    CUB~\cite{WahCUB_200_2011}             & \textcolor{customgreen}{\ding{51}}    & $1.5$K      & $100$       & $4.5$K      & $200$ \\
    Stanford-Cars~\cite{stanfordCars}   & \textcolor{customgreen}{\ding{51}}    & $2.0$K      & $98$        & $6.1$K      & $196$ \\    
    FGVC-Aircraft~\cite{aircrafts}   & \textcolor{customgreen}{\ding{51}}    & $1.7$K      & $50$        & $5.0$K      & $100$ \\
    \bottomrule
    \end{tabular}
    \vspace{-15pt}
\end{table}

%% file: tabs/benchmark_ncd.tex
\begin{table*}[t]

\centering

\caption{Comparison with state-of-the-art NCD methods. Results are reported in $ACC$ using task-aware evaluation protocol.
}

\renewcommand\arraystretch{1}

\setlength{\tabcolsep}{1.3mm}

\resizebox{0.75\linewidth}{!}{

\begin{tabular}{ll||ccccccccc@{}}

\hline
\rowcolor[HTML]{D3D3D3}
&Method & CUB & Stanford-Cars & FGVC-Aircraft & CIFAR-10 & CIFAR100-20 & CIFAR100-50 & ImageNet-100 & ImageNet-1K\\ 

\hline \hline

\multirow{5}{*}{\rotatebox{90}{$\leq$ 2020}}

&KCL~\cite{Hsu18_L2C} & - & - & - & 72.3$\pm$0.2 & 42.1$\pm$1.8 & - & - &73.8\\

&MCL~\cite{Hsu19_MCL} & - & - & - & 70.9$\pm$0.1 & 21.5$\pm$2.3 & - & - &74.4\\

&DTC~\cite{Han2019DTC} & - & - & - & 88.7$\pm$0.3 & 67.3$\pm$1.2 & 35.9$\pm$1.0 & - &78.3\\

&RS~\cite{han2019automatically} &-&-&- &90.4$\pm$0.5 &73.2$\pm$2.1 &39.2$\pm$2.3 &- &82.5\\

&RS+~\cite{han2019automatically} & 55.3$\pm$0.8$^\dagger$ & 36.5$\pm$0.6$^\dagger$ & 38.4$\pm$0.6$^\dagger$ & 91.7$\pm$0.9 & 75.2$\pm$4.2 & 44.1$\pm$3.7 & - &82.5\\

\hline

\multirow{5}{*}{\rotatebox{90}{2021}}
&Qing~\etal ~\cite{QING202124} & - & - & - &93.6$\pm$0.6  &81.3$\pm$1.9  & - &- &- \\
&OpenMix~\cite{openmix2020} & - & - & - & 95.3 & 87.2 & - & 74.76 &85.7\\
&NCL~\cite{ncl} & 48.1$\pm$0.9$^\dagger$ & 43.5$\pm$1.2$^\dagger$ & 43.0$\pm$0.5$^\dagger$ & 93.4$\pm$0.5 & 86.6$\pm$0.4 & 52.7$\pm$1.2 & - &90.7\\

&JOINT~\cite{Jia2021JointRL} & - & - & - & 93.4$\pm$0.6 & 76.4$\pm$2.8 & - & - &86.7\\

&DualRS~\cite{dualRs} & - & - & - & 91.6$\pm$0.6 & 75.3$\pm$2.3 & - & - &88.9\\

&UNO~\cite{uno} & 59.2$\pm$0.4$^\dagger$ & 49.8$\pm$1.4$^\dagger$ & 52.1$\pm$0.7$^\dagger$ & 93.3$\pm$0.4 & 90.5$\pm$0.7 & 62.3$\pm$1.4 & 79.56 &90.7\\

\hline

\multirow{6}{*}{\rotatebox{90}{$\geq$ 2022}}
&ResTune~\cite{ResTune} &- &- &- &89.0$\pm$0.6 &63.7$\pm$1.0 &- &- &-\\

&IIC~\cite{Li2023ncdiic} & 71.3$\pm$0.6 & 55.2$\pm$0.7 & 56.0$\pm$0.8 & 99.1$\pm$0.0 & 92.4$\pm$0.2 & 65.8$\pm$0.9 & 80.24 &91.9\\

&CRKD~\cite{peiyan2023class} & 65.7$\pm$0.6 & 53.5$\pm$0.8 & 55.8$\pm$0.9 & 93.5$\pm$0.3 & 91.2$\pm$0.1 & 65.3$\pm$0.6 & 80.94 &-\\

&NSCL~\cite{sun2023nscl} & -& -& - &97.5&85.9&64.0& - &-\\

&APL~\cite{10328468}  & -& -& - &96.0$\pm$0.4&78.9$\pm$0.9& -& - &88.8\\

&SCKD~\cite{wang2024selfcooperationknowledgedistillationnovel} & 73.1$\pm$0.4 & 56.8$\pm$0.7 & 56.5$\pm$0.7 & 95.6$\pm$0.2 & 92.6$\pm$0.6 & 68.2$\pm$0.4 & 82.18 &-\\ 

\bottomrule

\end{tabular}

}

\label{tab:ncd_results}
\RaggedRight
~~~~~~~~~~~~~~~~~~~~~~~\footnotesize{$^\dagger$ denotes results from~\cite{peiyan2023class}.}
\vspace{-15pt}
\end{table*}

%% file: tabs/benchmark_gcd.tex
\begin{table*}[ht]
\vspace{-2pt}
  \caption{Comparison of state-of-the-art GCD methods. Results are reported in $ACC$ across `All', `Old' and `New' categories. }
  \vspace{-2pt}
  \label{tab:gcd_results}
  \centering
  \setlength{\tabcolsep}{1.3mm}
  \resizebox{1.0\linewidth}{!}{
\begin{tabular}{ll||ccc|ccc|ccc|ccc|ccc|ccc}
\hline
\rowcolor[HTML]{D3D3D3}
&&\multicolumn{3}{c|}{CUB}& \multicolumn{3}{c|}{Stanford-Cars}&\multicolumn{3}{c|}{FGVC-Aircraft}&\multicolumn{3}{c|}{CIFAR-10}& \multicolumn{3}{c|}{CIFAR-100}&\multicolumn{3}{c}{ImageNet-100}\\
\rowcolor[HTML]{D3D3D3}&Method
&All&Old&New&All&Old&New&All&Old&New&All&Old&New &All&Old&New &All&Old&New\\
\hline \hline
\multirow{4}{*}{\rotatebox{90}{2022}}
&RankStats+~\cite{han2021autonovel}&33.3&51.6&24.2 &28.3&61.8&12.1 &26.9&36.4&22.2 &46.8&19.2&60.5 &58.2&77.6&19.3 &37.1&61.6&24.8\\
&UNO+~\cite{uno}&35.1&49.0&28.1 &35.5&70.5&18.6 &40.3&56.4&32.2 &68.6& {98.3}&53.8 &69.5&80.6&47.2 &70.3&95.0&57.9\\
&XCon~\cite{fei2022xcon}&52.1&54.3&51.0 &40.5&58.8&31.7 &47.7&44.4&49.4 &96.0&97.3&95.4 &74.2&81.2&60.3 &77.6&93.5&69.7\\
&GCD~\cite{vaze2022gcd}&51.3&56.6&48.7 &39.0&57.6&29.9 &45.0&41.1&46.9 &91.5&{97.9}&88.2 &73.0&76.2&66.5 &74.1&89.8&66.3\\
\hline
\multirow{8}{*}{\rotatebox{90}{2023}}
&SimGCD~\cite{wen2023simgcd}&60.3&65.6&57.7 &53.8&71.9&45.0 &54.2&59.1&51.8 &97.1&95.1&98.1 &80.1&81.2&77.8 &83.0&93.1&77.9\\
&PromptCAL~\cite{zhang2022promptcal}&62.9&64.4&62.1 &50.2&70.1&40.6 &52.2&52.2&52.3 & {97.9}&96.6&  {98.5} &81.2&84.2&75.3 &83.1&92.7&78.3\\
&DCCL~\cite{pu2023dynamic}&63.5&60.8&64.9 &43.1&55.7&36.2 &42.9$^\dagger$&44.5$^\dagger$&42.1$^\dagger$ &96.3&96.5&96.9 &75.3&76.8&70.2 &80.5&90.5&76.2\\
&GPC~\cite{Zhao_2023_ICCV}&52.0&55.5&47.5 &38.2&58.9&27.4 &43.3&40.7&44.8 &90.6&97.6&87.0 &75.4&{84.6}&60.1 &75.3&93.4&66.7\\
&PIM~\cite{chiaroni2023parametric}&62.7&75.7&56.2 &43.1&66.9&31.6 &48.8$^\dagger$&60.0$^\dagger$&43.0$^\dagger$ &94.7&97.4&93.3 &78.3&84.2&66.5 &83.1&  \underline{95.3}&77.0\\
&$\mu$GCD$^\ddagger$~\cite{vaze2023clevr4} &65.7&68.0&64.6 &56.5&68.1&50.9 &53.8&55.4&53.0 &-&-&- &-&-&- &-&-&-\\
&InfoSieve~\cite{rastegar2023learn}&69.4& {77.9}&65.2 &55.7&74.8&46.4 &56.3&63.7&52.5 &94.8&97.7&93.4 &78.3&82.2&70.5 &80.5&93.8&73.8\\
&CAC$^\ddagger$~\cite{yang2023GCDwithclustering}&58.0& 65.0&43.9 &47.6&70.6&33.8 &-&-&- &92.3&91.4&94.4 &78.5&81.4&75.6 &81.1&80.3&81.8\\
\hline
\multirow{25}{*}{\rotatebox{90}{$\geq 2024$}}
&CiPR~\cite{hao2023cipr} &57.1&58.7&55.6 &47.0&61.5&40.1 &44.0$^\dagger$&50.5$^\dagger$&40.8$^\dagger$ 
&  {97.7}&97.5&{97.7} &{81.5}&82.4&{79.7} &80.5&84.9&78.3\\
&SPTNet~\cite{wang2024sptnet} &65.8&68.8&65.1 &59.0&{79.2}&49.3 &{59.3}&61.8&58.1
&{97.3}&95.0& {98.6} &81.3&84.3&75.6 &{85.4}&93.2&{81.4}\\
&LegoGCD~\cite{Cao_2024_CVPR} &63.8&71.9&59.8 &57.3&{75.7}&48.4 &{55.0}&61.5&51.7
&{97.1}&94.3& {98.5} &81.8&81.4&82.5 &86.3&94.5&{82.1}\\
&CMS~\cite{choi2024contrastive}&68.2&  {76.5}&64.0 &56.9&76.1&47.6 &56.0&63.4&52.3 
&95.7$^\dagger$&97.0 $^\dagger$&94.4$^\dagger$&  {82.3}& \underline{85.7}&75.5 &84.7& \bf{95.6}&79.2\\
&Yang~\etal$^\ddagger$~\cite{yang2024learning} &61.3&60.8&62.1 &44.3&58.2&39.1 &-& -&- &96.5&97.6&94.4 &74.6&76.5& {69.4} &78.1&91.3&70.5\\
&AMEND~\cite{Banerjee_2024_WACV} &64.9&75.6&59.6 &52.8&61.8&48.3 &56.4& \bf{73.3}&48.2 &96.8&94.6&97.8 &81.0&79.9& \underline{83.3} &83.2&92.9&78.3\\
&GCA~\cite{Otholt_2024_WACV} &68.8&73.4&66.6 &54.4&72.1&45.8 &52.0& 57.1&49.5 &95.5&95.9&95.2 &82.4&{85.6}& 75.9 &82.8&94.1&77.1\\
&SelEx~\cite{RastegarECCV2024}&  {73.6}&75.3&{72.8} &58.5&75.6&50.3 &57.1&64.7&53.3 
&95.9&\bf{98.1}&94.8 &{82.3}&{85.3}&76.3 &83.1&93.6&77.8\\
&MSGCD~\cite{DUAN2025103020}&  {63.6}&70.7&{60.0} &57.7&75.5&49.9 &56.4&64.1&52.6 
&97.1$^\dagger$&97.4$^\dagger$&97.0$^\dagger$&{81.4}&{82.5}&79.0 &82.1$^\dagger$&88.9$^\dagger$&78.7$^\dagger$\\
&FlipClass~\cite{lin2024flipped}&  {71.3}&71.3&{71.3} &63.1&\underline{81.7}&53.8 &59.3&{66.9}&55.4 
&\bf{98.5}&{97.6}&\bf{99.0} &\bf{85.2}&{84.9}&\bf{85.8} &{86.7}&94.3&\bf{82.9}\\
&OwMatch~\cite{niu2024owmatch}&  {70.5}$^\dagger$&\bf{81.5}$^\dagger$&{65.0}$^\dagger$ &58.0$^\dagger$&75.5$^\dagger$&49.3$^\dagger $&52.6$^\dagger$&63.2$^\dagger$&47.3$^\dagger$ 
&96.6$^\dagger$&\underline{98.0}$^\dagger$&95.9$^\dagger$ &{79.8}$^\dagger$&{81.2}$^\dagger$&79.1$^\dagger$ &84.7$^\dagger$&\bf{95.6}$^\dagger$&79.2$^\dagger$\\
&DebGCD~\cite{liu2025dg}&{66.3}&71.8&{63.5} &65.3&81.6&57.4 &\underline{61.7}&63.9&{60.6} 
&97.2&94.8&98.4 &{83.0}&{84.6}&79.9 &85.9&94.3&81.6\\
&Contextuality-GCD~\cite{luo2024contextualityhelpsrepresentationlearning}&64.5&68.7&61.5 &55.8&73.1&47.8 &56.8&61.8&54.4 &97.3&96.1&\underline{98.8}	&82.0&83.9&78.9 &84.6&94.0&81.0\\
&MOS~\cite{peng2025mos}&69.6&72.3&68.2&64.6&80.9&56.7&61.1&66.9&58.2&-&-&-&-&-&-&-&-&- \\
&AptGCD~\cite{zhang2025less}&70.3&74.3&69.2&62.1&79.7&53.6&61.1&65.2&59.0&97.3&95.8&98.7&82.8&81.8&85.5&\bf{87.8}&95.4&84.3 \\
&Dai~\etal~\cite{dai2025adaptive}&68.5&73.1&66.2&62.3&80.7&53.4&60.9&63.5&59.6&-&-&-&-&-&-&-&-&-\\
&HypCD~\cite{liu2025hyperbolic}&\underline{79.8}&75.8&\underline{81.8} &{62.9}&{80.0}&{54.7} &\bf{65.9}&\underline{67.3}&\bf{65.1}
&96.7&97.6&96.3 
&{82.4}&85.1&77.0
&\underline{86.8}&94.6&\underline{82.8}\\
&SEAL~\cite{he2025seal}&66.2&{72.1}&63.2&{65.3}&{79.3}&{58.5}&{62.0}&{65.3}&{60.4}&97.2&94.7&98.4 &{82.1}&{81.7}&{83.0} &84.6&90.9&81.3\\
&PAL-GCD$^\ddagger$~\cite{wang2025prior}&  {72.6}&75.2&{71.2} &\bf{72.2}&\bf{83.4}&\bf{66.8} &59.5&{67.2}&55.6 
&-&-&- &{82.0}&{82.4}&81.2 &86.3&93.0&82.0\\
&ConGCD~\cite{tang2025dissecting}&\bf{81.7}&\underline{80.4}&\bf{82.4} &57.5&77.5&47.9 &62.5&70.2&58.7 
&97.4&95.2&98.5 &{82.5}&\bf{85.9}&77.3 &{85.9}&93.4&82.5\\
&AF~\cite{xu2025hidden}&{69.0}&{74.3}&{66.3} &67.0&80.7&60.4 &59.4&68.1&55.0 
&\underline{97.8}&95.9&\underline{98.8} &{82.2}&{85.0}&76.5 &{85.4}&94.6&80.8\\
&PNP$^\ddagger$~\cite{wang2024knownclustersprobenew}&67.4&69.2&66.5 &57.3&70.2&51.1 &47.0&49.5&45.7 &90.2&95.0&87.8 &80.0&81.7&76.3 &82.4&92.9&77.2\\
&RPIM$^\ddagger$~\cite{tan2024revisitingmutualinformationmaximization}&66.8&77.3&61.5 &45.8&67.5&35.3 &-&-&- &95.3&97.6&94.1 &82.7&85.4&77.2 &83.2&{95.2}&77.2\\
&ConceptGCD$^\ddagger$~\cite{zhang2024composing}&69.4&75.4&66.5 &\underline{70.1}&81.6&\underline{64.6} &{60.5}&59.2&\underline{61.1} &-&-&- &82.8&84.1&80.1 &86.3&93.3&\underline{82.8}\\
&MCDL$^\ddagger$~\cite{tu2024memoryconsistencyguideddivideandconquer}&68.7&74.2&63.4 &61.9&77.9&54.7 &57.6&64.6&52.5 &97.3&95.4&98.4 &\underline{84.0}&84.7&81.2 &83.8&93.8&78.8\\
&NN-GCD$^\ddagger$~\cite{ji2024fresh}&72.4&76.3&70.5 &65.4&77.5&59.5 &{60.5}&65.3&58.2 &97.3&95.0&98.6 &81.2&82.4&78.8 &85.2&91.5&82.2\\
\hline
\end{tabular}
}
\RaggedRight
\noindent\footnotesize{$^\dagger$ denotes results from our implementation.}
\noindent\footnotesize{$^\ddagger$ denotes methods without releasing official codes.}
\vspace{-16pt}
\end{table*}



  

%% file: tabs/benchmark_gcd_dinov2.tex
\begin{table*}[ht]
\vspace{-1pt}
  \caption{Additional Comparison of state-of-the-art GCD methods using ViT backbone pretrained by DINOv2~\cite{oquabdinov2} and CLIP~\cite{clip}. }
  \vspace{-1pt}
  \label{tab:gcd_results_dino2}
  \centering
  \setlength{\tabcolsep}{1.3mm}{
  \resizebox{1.0\linewidth}{!}{
\begin{tabular}{ll||ccc|ccc|ccc|ccc|ccc|ccc}
\hline 
\rowcolor[HTML]{D3D3D3}&&\multicolumn{3}{c|}{CUB}& \multicolumn{3}{c|}{Stanford-Cars}&\multicolumn{3}{c|}{FGVC-Aircraft}&\multicolumn{3}{c|}{CIFAR-10}& \multicolumn{3}{c|}{CIFAR-100}&\multicolumn{3}{c}{ImageNet-100}\\ 
\rowcolor[HTML]{D3D3D3}&Method
&All&Old&New&All&Old&New&All&Old&New&All&Old&New &All&Old&New &All&Old&New\\
\hline \hline
\multirow{9}{*}{\rotatebox{90}{\emph{DINOv2}}}
&$\mu$GCD$^\ddagger$~\cite{vaze2023clevr4}&74.0&75.9&73.1 &{76.1}&{91.0}&68.9 &66.3&68.7&65.1 &-&-&- &-&-&- &-&-&-\\
&CiPR~\cite{hao2023cipr}&{78.3}&73.4&{80.8} &66.7&77.0&61.8 &59.1$^\dagger$&56.9$^\dagger$&60.2$^\dagger$ &\bf{99.0}&{98.7}&99.2 &\bf{90.3}&89.0&\bf{93.1} &88.2&87.6&{88.5}\\
&SPTNet~\cite{wang2024sptnet} &76.3&79.5&74.6 &72.3$^\dagger$&82.0$^\dagger$&67.5$^\dagger$ &68.0$^\dagger$&75.2$^\dagger$&60.8$^\dagger$ &\underline{98.9}$^\dagger$&\bf{99.1}$^\dagger$ &98.8$^\dagger$&89.0$^\dagger$&\bf{91.5}$^\dagger$&79.2$^\dagger$ &{90.1}&{96.1}&87.1\\
&GCD~\cite{vaze2022gcd}&71.9$^\triangle$&71.2$^\triangle$&72.3$^\triangle$ &65.7$^\triangle$&67.8$^\triangle$&64.7$^\triangle$ &55.4$^\triangle$&47.9$^\triangle$&59.2$^\triangle$ &97.8$^\diamond$&\underline{99.0}$^\diamond$&97.1$^\diamond$ &79.6$^\diamond$&84.5$^\diamond$&69.9$^\diamond$ &78.5$^\diamond$&89.5$^\diamond$&73.0$^\diamond$\\
&SimGCD~\cite{wen2023simgcd}&71.5$^\triangle$&78.1$^\triangle$&68.3$^\triangle$ &71.5$^\triangle$&81.9$^\triangle$&66.6$^\triangle$ &63.9$^\triangle$&69.9$^\triangle$&60.9$^\triangle$ &98.7$^\diamond$&96.7$^\diamond$&\bf{99.7}$^\diamond$ &88.5$^\diamond$&{89.2}$^\diamond$&{87.2}$^\diamond$ &89.9$^\diamond$&95.5$^\diamond$&87.1$^\diamond$\\
&SelEx~\cite{RastegarECCV2024} &\underline{87.4}&\underline{85.1}&\underline{88.5} &\underline{82.2}&\bf{93.7}&\underline{76.7} &\underline{79.8}&\bf{82.3}&\underline{78.6} 
&98.5$^\dagger$&98.8$^\dagger$&98.5$^\dagger$ &87.7$^\dagger$&{90.8}$^\dagger$&81.5$^\dagger$ &{90.9}$^\dagger$&{96.2}$^\dagger$&88.3$^\dagger$\\
&DebGCD~\cite{liu2025dg}&77.5&{80.8}&75.8 &75.4&87.7&{69.5} &{71.9}&{76.0}&{69.8} &\underline{98.9}&97.5&\underline{99.6} &\underline{90.1}&\underline{90.9}&{88.6} &\bf{93.2}&\bf{97.0}&\bf{91.2}\\
&Dai~\etal~\cite{dai2025adaptive}&75.1&79.1&73.2&73.4&87.6&66.7&68.8&74.1&66.6&-&-&-&-&-&-&-&-&-\\
&HypCD~\cite{liu2025hyperbolic}&\bf{90.7}&\bf{85.3}&\bf{93.4} &\bf{83.8}&\underline{93.3}&\bf{79.2} &\bf{83.4}&\underline{82.0}&\bf{84.1} 
&98.6&98.1&98.9 
&88.6&\bf{91.5}&82.8
&\underline{92.3}&\underline{96.4}&{90.2}\\
&SEAL~\cite{he2025seal}&76.7&78.3&{75.9}&{77.7}&{88.7}&{72.4}&{74.6}&{73.2}&{75.3}&\underline{98.9}&98.1&{99.3} &{89.8}&{90.4}&\underline{89.5} &{91.3}&93.3&\underline{90.3}  \\
\hline
\multirow{6}{*}{\rotatebox{90}{\emph{CLIP}}}
&GCD~\cite{vaze2022gcd}&57.6$^\circ$&65.2$^\circ$&53.8$^\circ$ &65.1$^\circ$&75.9$^\circ$&59.8$^\circ$ &45.3$^\circ$&44.4$^\circ$&45.8$^\circ$&94.0$^\circ$&\underline{97.3}$^\circ$&92.3$^\circ$ &74.8$^\circ$&79.8$^\circ$&64.6$^\circ$ &75.8$^\circ$&87.3$^\circ$&70.0$^\circ$\\
&SimGCD~\cite{wen2023simgcd}&71.7$^\circ$&76.5$^\circ$&69.4$^\circ$ &70.0$^\circ$&83.4$^\circ$&63.5$^\circ$ &\underline{54.3}$^\circ$&\underline{58.4}$^\circ$&\underline{52.2}$^\circ$ &97.0$^\circ$&94.2$^\circ$&98.4$^\circ$ &81.1$^\circ$&85.0$^\circ$&73.3$^\circ$ &\underline{90.8}$^\circ$&\underline{95.5}$^\circ$&\underline{88.5}$^\circ$\\
&TextGCD~\cite{zheng2024textualknowledgematterscrossmodality}&\underline{76.6}&\bf{80.6}&\underline{74.7} &\bf{86.9}&\underline{87.4}&\bf{86.7} &-&-&-&\bf{98.2}&\bf{98.0}&\bf{98.6} &\bf{85.7}&\bf{86.3}&\underline{84.6} &88.0&92.4&85.2\\
&GET$^\ddagger$~\cite{yang2025consistent}&\bf{77.0}&\underline{78.1}&\bf{76.4} &\underline{78.5}&86.8&\underline{74.5} &\bf{58.9}&\bf{59.6}&\bf{58.5}&\underline{97.2}&94.6&\underline{98.5} &82.1&\underline{85.5}&75.5 &\bf{91.7}&\bf{95.7}&\bf{89.7}\\
&CLIP-GCD$^\ddagger$~\cite{ouldnoughi2023clipgcdsimplelanguageguided}&62.8&77.1&55.7 &70.6&\bf{88.2}&62.2 &50.0&56.6&46.5&96.6&97.2&96.4 &\underline{85.2}&85.0&\bf{85.6} &84.0&\underline{95.5}&78.2\\
&CPT$^\ddagger$~\cite{wang2024unlockingmultimodalpotentialclip}&70.1&73.5&68.4 &74.2&84.3&69.3 &-&-&-&96.3&96.2&96.4 &81.3&{81.3}&{81.2} &89.2&95.2&86.1\\
\hline
\end{tabular}
}
}
\RaggedRight
\noindent\footnotesize{$^\circ, ^\triangle$ and $^\diamond$ denote results from \cite{wang2024unlockingmultimodalpotentialclip,vaze2023clevr4} and \cite{hao2023cipr} respectively.}
\noindent\footnotesize{$^\ddagger$ denotes methods without releasing official codes.}
\noindent\footnotesize{$^\dagger$ denotes results implemented by us.}
\vspace{-15pt}
\end{table*}

%% file: Secs/5_Future.tex
\vspace{-10pt}

\section{Discussion and Forward Looking}
\label{Sec:future}

\noindent
\textbf{Label Assignment: Parametric \textit{vs.} Non-Parametric.}  
{SimGCD~\cite{wen2023simgcd} introduces a parametric classifier for GCD, demonstrating strong performance. However, subsequent works such as SelEx~\cite{RastegarECCV2024} and CMS~\cite{choi2024contrastive} have shown consistent improvements with non-parametric clustering–based approaches, highlighting their flexibility in capturing complex category structures. 
These contrasting findings suggest that there is likely no universally optimal label assignment strategy for CD. Instead, the effectiveness of label assignment appears to be highly dependent on the representation learning paradigm and even the dataset characteristics (\eg, class granularity, distribution skew). 
{Thus, it is worth studying} adaptive or hybrid assignment strategies that better align with the learned representations and dataset characteristics, rather than relying on a one-size-fits-all solution.}

\noindent
\textbf{Efficient and Effective Class Number Estimation.}  
{
A key challenge in category discovery is determining the number of novel classes. Most existing approaches~\cite{vaze2022gcd, wen2023simgcd} treat estimation as a post-hoc process, performed after representation learning is complete. 
%
{More recently, methods such as GPC~\cite{Zhao_2023_ICCV} and CiPR~\cite{hao2023cipr} have explored jointly estimating class numbers during representation learning, showing that the two processes can mutually reinforce each other.
Nevertheless, open questions remain around efficiency, scalability, and stability in complex scenarios. This calls for future research in} developing frameworks that can adaptively infer class numbers during training in an effective and efficient manner.
}

\noindent
\textbf{Cross-Modal Information.}  
{With the advancement of multimodal learning, leveraging complementary modalities, particularly text, has proven effective in various visual tasks~\cite{zhang2024multimodal, clip}. 
{
In the CD literature, research has mainly focused on visual-only pipelines, leaving the role of multimodal cues comparatively underexplored. This raises the question of how vision–language models (VLMs) might support category discovery.} Models like CLIP~\cite{clip}, pretrained on large-scale image-text pairs, offer rich, semantically grounded representations that can provide strong initialization for category discovery. Yet, this advantage comes with concerns: recent findings~\cite{ mayilvahanan2024search} suggest its open-world performance may stem from memorization rather than generalization. This raises the risk of data leakage, especially when evaluation datasets overlap the CLIP training data. To ensure fair benchmarking, it is recommended to \textit{include results on datasets disjoint from CLIP's training data or across diverse domains with varying overlap} such as the use of NEV in GET~\cite{wang2024unlockingmultimodalpotentialclip}, which lies outside CLIP's pretraining.}

{Empirical comparisons in Tab.~\ref{tab:gcd_results} and Tab.~\ref{tab:gcd_results_dino2} also show that self-supervised vision models like DINOv2~\cite{oquabdinov2} can outperform CLIP~\cite{clip} without textual information, suggesting that representations learned purely from visual structures may be better suited for unsupervised category separation. 
These observations do not rule out the utility of VLMs in category discovery; rather, they highlight the importance of how and when to leverage multimodal signals.
{Going forward, an interesting direction lies in explicitly incorporating semantic priors extracted from language}, such as textual descriptions or attribute names, without overly relying on VLM backbones. This may offer the benefits of multimodal supervision while avoiding potential pitfalls tied to memorization or domain bias.}

\noindent
\textbf{From Single-Object to Complex Multi-Object Scenes.}  
Most CD studies have been developed and evaluated on single-object datasets, a design convenient for benchmarking but unrepresentative of real scenes that typically contain multiple interacting objects, cluttered backgrounds, and occlusions.
{Consequently, most GCD methods implicitly operate at the image level under a single-salient-object assumption, limiting their ability to localize and disentangle multiple co-occurring instances.} Extending CD to complex scenes introduces additional questions: how to disentangle features across instances, which objects should be discovered (potentially conditioned on user interest), and how to associate instance-level evidence with emerging categories.
{These considerations motivate advancing CD toward realistic multi-object environments.}

\noindent
\textbf{Toward More Real-world Settings.}
{In real-world deployments, CD systems are often required to operate continually, adapt to new data over time, and handle class imbalance and domain shifts simultaneously. }This calls for unified frameworks that can integrate continual learning, cross-domain generalization,  distribution-agnostic modeling, and more. For instance, continual category discovery under imbalanced and heterogeneous domains remains largely underexplored, yet it closely mirrors real-world challenges such as dynamic environments in robotics or evolving data in medical diagnostics.
{Beyond image-level recognition, it is natural to consider extensions to instance- or pixel-level discovery in multi-object scenes, as well as to related tasks such as 3D depth estimation, video analysis, and action recognition. }

%% file: Secs/6_conclusion.tex
\vspace{-10pt}

\section{Conclusion}
\label{Sec:conclusion}
This survey provides a comprehensive review of the latest developments in the open-world problem of \textit{Category Discovery}, marking the first extensive literature review on this topic.
We begin by providing clear conceptual definitions for category discovery and systematically comparing them with related research areas.
Existing methods are then organized by base settings and derived settings, followed by an in-depth analysis of their core components: \textit{representation learning}, \textit{label assignment}, and \textit{category number estimation}.
We also provide a comparative benchmark using commonly used datasets, offering a quantitative evaluation for all the methods.
Finally, we distill key takeaways from prior studies and outline concrete, forward-looking research directions.

%% file: Secs/appendix.tex
\setcounter{table}{0} 
\setcounter{figure}{0} 
\setcounter{page}{1}
\renewcommand{\thetable}{A\arabic{table}}  
\noindent
\textbf{Overview.} In this appendix, we provide an off-the-shelf benchmark for class number estimation, offering a unified comparison of existing methods. Additionally, we include further experimental details and results, as well as extended descriptions of the datasets commonly used in category discovery tasks.

\section{Benchmark of Estimating the number of unknown categories}

\begin{table*}[h]
\caption{Estimating the number of classes \(K\) and the error rate (Err\%).}
\label{table:ablation_kestimation}
\centering
    \resizebox{0.7\linewidth}{!}{%
    \tabcolsep=0.12cm
    \begin{tabular}[b]{ll |cc| cc| cc| cc| cc| cc}
    \hline
     \rowcolor[HTML]{D3D3D3} &  {Method} &
      \multicolumn{2}{c|}{CIFAR10} &
        \multicolumn{2}{c|}{CIFAR100} &
        \multicolumn{2}{c|}{ImageNet-100} &
        \multicolumn{2}{c|}{CUB} &
        \multicolumn{2}{c|}{Stanford Cars} &
        \multicolumn{2}{c}{FGVC Aircraft} \\
   \rowcolor[HTML]{D3D3D3}      && K & Err(\%) &  K & Err(\%) & K & Err(\%)  & K & Err(\%) &  K & Err(\%) & K & Err(\%) \\
    \hline \hline
    &Ground truth & 10 & - & 100 & - & 100 & - & 200 & - & 196 & - & 100 & -  \\
    \hline
    \multirow{7}{*}{\rotatebox{0}{\emph{DINOv1}}}
        
        &GCD~\cite{vaze2022gcd} &9 & 1& 100 & 0 & 109 & 9  & 231 & 15.5 & 230 & 17.3 & - & -   \\
        &DCCL~\cite{pu2023dynamic} & 14&4&146 & 46 & 129 & 29& 172 & 14 & 192 & 0.02 & - & - \\
        &PIM~\cite{chiaroni2023parametric} &10&0& 95 & 5 & 102 & 2 & 227 & 13.5 & 169 & 13.8 & - & -  \\
        &CiPR~\cite{hao2023cipr} & 12&2&103 & 3 & 100 & 0& 182 & 9 & 182 & 7.14 & - & - \\
        &GPC~\cite{Zhao_2023_ICCV} & 10&0&100 & 0 & 103 & 3 & 212 & 6 & 201 & 0.03 & - & - \\
        &PAL-GCD~\cite{wang2025prior} & -&-&- & - & - & - & 176 & 12 & 172 & 12.2 & 77 & 23 \\
        &CMS~\cite{choi2024contrastive}&-&- & 97 & 3 & 116 & 16 & 170 & 15 & 156 & 20.4 & 98 & 2  \\
    \hline
    \multirow{2}{*}{\rotatebox{0}{\emph{DINOv2}}}
    &GCD~\cite{vaze2022gcd} & 10&0 &100  &0  & 100 & 0  &  211&5.5  & 215 & 9.69 & 105 & 5   \\
    &Perez \etal$^\ddagger$~\cite{perez2024human}  & 10&0 & 100 & 0 &  101& 1  & 202 & 1 & 194 & 1.02 & 123 & 23   \\
    \hline
    \multicolumn{14}{l}{\small $^\ddagger$ denotes our implementation using the DINOv2 pretrained ViT-Base backbone.}
    \end{tabular}%
}
\end{table*}%

{
We further report the estimated number of categories for six commonly used datasets. 
Table~\ref{table:ablation_kestimation} presents the estimated number of classes (K) and the corresponding error rates (Err\%) across various datasets for several methods, including GCD~\cite{vaze2022gcd}, DCCL~\cite{pu2023dynamic}, PIM~\cite{chiaroni2023parametric}, CiPR~\cite{hao2023cipr}, GPC~\cite{Zhao_2023_ICCV}, PAL-GCD~\cite{wang2025prior}, and CMS~\cite{choi2024contrastive}. The ground truth number of classes for each dataset (CIFAR-10~\cite{krizhevsky2009cifar}, CIFAR-100~\cite{krizhevsky2009cifar}, ImageNet-100~\cite{russakovsky2015imagenet}, CUB~\cite{WahCUB_200_2011}, Stanford-Cars~\cite{stanfordCars}, and FGVC-Aircraft~\cite{aircrafts}) is provided in the first row for reference. }

{
Beyond methods developed within the GCD community, other domains have also proposed effective strategies for estimating the number of categories in unlabelled datasets. For example, Perez~\etal~\cite{perez2024human} introduce a class number estimation approach for Re-ID tasks based on nested importance sampling and human-in-the-loop feedback. Their method involves limited human supervision to determine whether a given image pair belongs to the same class, which aligns well with our partially labelled setting. To evaluate its applicability in GCD, we reimplement their method using a DINOv2\cite{oquabdinov2} ViT-Base backbone and apply it to both fine-grained and generic datasets. For fine-grained datasets, we sample only $0.1\%$ of all possible image pairs from the labelled data, while for CIFAR-10 and CIFAR-100, we use just $0.01\%$.  For ImageNet-100, due to its larger image count, we only employ $5000$ pairs, which is around $10^{-6}$ of all possible image pairs from the labelled data.  The estimation results are reported in Table~\ref{table:ablation_kestimation}.
}

\input{tabs/CCD_dataset_statistics}

\input{tabs/benchmark_ccd}

\section{Benchmark of Continual Category Discovery (CCD)}
\label{Subsec:CCD_benchmark}
We additionally report the benchmark results for CCD methods on fine-grained datasets, including CUB~\cite{WahCUB_200_2011}, Stanford-Cars~\cite{stanfordCars} and FGVC-Aircraft~\cite{aircrafts} as well as generic datasets including CIFAR-100~\cite{krizhevsky2009cifar}, ImageNet-100~\cite{russakovsky2015imagenet}, TinyImageNet~\cite{le2015tinyImagenet} and Caltech-101~\cite{fei2006one} based on DINO~\cite{caron2021dino} pre-trained ViT-Base~\cite{dosovitskiy2020image} backbone.
Table~\ref{tab:ccd_results} shows the performance of seven state-of-the-art GCD methods. 
The performance is measured by $cACC$ across the `All', `Old' and `New' categories, as introduced in Section IV (Evaluation Protocol). 

Notably, we employ the data splitting strategy in PromptCCD~\cite{cendra2024promptccd} with details reported in Table~\ref{tab:CCD_data_dist}.

\section{Additional Experimental Details of GCD Benchmark}
In this section, we provide results on additional datasets - Oxford-Pets~\cite{oxfordPets}, Herbarium19~\cite{Chuan19herbarium} and ImageNet-$1$K~\cite{russakovsky2015imagenet}. The performance is measured by $ACC$ across the `All', `Old' and `New' categories. Table~\ref{tab:pets} shows the performance of $12$ state-of-the-art GCD methods on this two datasets based on DINO~\cite{caron2021dino} pre-trained ViT-B~\cite{dosovitskiy2020image} backbone. 

\begin{table}[h!]
\centering
\caption{Additional experiments comparison for GCD methods on Herbarium19~\cite{Chuan19herbarium} and Oxford-Pets~\cite{oxfordPets}.} 
\setlength{\tabcolsep}{3mm}{
\resizebox{0.45\textwidth}{!}{
\begin{tabular}{l|ccc|ccc}
    \hline
    \rowcolor[HTML]{D3D3D3} &\multicolumn{3}{c|}{Oxford-Pets~\cite{oxfordPets}}&\multicolumn{3}{c}{Herbarium19~\cite{Chuan19herbarium}}\\
\rowcolor[HTML]{D3D3D3} Method&All&Old&New&All&Old&New\\ 
    \hline \hline
    ORCA~\cite{orca2022}&-&-&- &24.6&26.5&23.7 \\
    GCD~\cite{vaze2022gcd}&80.2&85.1&77.6 &35.4&51.0&27.0\\
    XCon~\cite{fei2022xcon}&86.7&91.5&84.1 &-&-&-\\
    OpenCon~\cite{sun2023opencon}&-&-&- &39.3&58.9&28.6\\
    DCCL~\cite{pu2023dynamic}&88.1&88.2&88.0 &-&-&- \\
    SimGCD~\cite{wen2023simgcd}&91.7&83.6&\underline{96.0} &44.0&58.0&36.4 \\
    $\mu$GCD~\cite{vaze2023clevr4}&-&-&- &\underline{45.8}&\textbf{61.9}&\underline{37.2} \\
    InfoSieve~\cite{rastegar2023learn}&90.7&\textbf{95.2}&88.4 &40.3&59.0&30.2 \\ 
    SelEx~\cite{RastegarECCV2024} &{92.5}&\underline{91.9}&92.8 &39.6&54.9&31.3 \\
    DebGCD~\cite{liu2025dg} &\bf{93.0}&86.4&\bf{96.5} &44.7&59.4&36.8 \\
    HypCD~\cite{liu2025hyperbolic}&{92.7}&91.5&93.3 &{45.1}&\underline{60.1}&{36.9}\\
    SEAL~\cite{he2025seal}&\underline{92.9}&88.9&95.0&\bf{46.9}&45.8&\bf{48.2}\\
    \hline
\end{tabular}
}
}
\label{tab:pets}
\end{table}

\begin{table}[h]
\centering
\caption{Additional experiments comparison for GCD methods on ImageNet-$1$K~\cite{russakovsky2015imagenet}.} 
\setlength{\tabcolsep}{3mm}{
\resizebox{0.4\textwidth}{!}{
\begin{tabular}{l|cccccc}
    \hline
    \rowcolor[HTML]{D3D3D3}  &\multicolumn{3}{c}{ImageNet-$1$K~\cite{russakovsky2015imagenet}}\\
 \rowcolor[HTML]{D3D3D3} Method&All&Old&New\\ 
    \hline \hline
    GCD~\cite{vaze2022gcd}&$52.5^\dagger$&$72.5^\dagger$&$42.2^\dagger$\\
    SimGCD~\cite{wen2023simgcd}&$57.1$&$77.3$&$46.9$\\
    Contextuality-GCD~\cite{luo2024contextualityhelpsrepresentationlearning}&$59.5$&$79.9$&$49.5$  \\
    LegoGCD~\cite{Cao_2024_CVPR} &$62.4$&$79.5$&$53.8$ \\
    DebGCD~\cite{liu2025dg} &$65.0$&$82.0$&$56.5$\\
    \hline

\multicolumn{4}{l}{\small $^\dagger$ denotes results from~\cite{wen2023simgcd}.}\\
\end{tabular}
}
}
\label{tab:add}
\end{table}

\begin{table*}[h]
\centering
\caption{Comparison of methods under different data splitting protocol. All methods below employ the ResNet~\cite{resnet} backbone.} 
\setlength{\tabcolsep}{3mm}{
\resizebox{0.7\textwidth}{!}{
\begin{tabular}{ll|ccc|ccc|ccc}
    \hline
   \rowcolor[HTML]{D3D3D3}  &&\multicolumn{3}{c|}{CIFAR10}& \multicolumn{3}{c|}{CIFAR100}&\multicolumn{3}{c}{ImageNet-100}\\
   \rowcolor[HTML]{D3D3D3}&Method&All&Old&New&All&Old&New&All&Old&New\\ 
    \hline \hline
    \multirow{3}{*}{\rotatebox{90}{\tiny{\makecell{Ind.\\$\mathcal{D_L}$\\$50\%$}}}}
    &OpenLDN~\cite{rizve2022openldn}&95.4&95.7&95.1 &59.3&74.1&44.5 &79.1&89.6&68.6\\
    &LPS~\cite{lps_ijcai2024}&92.4&90.2&93.4 &54.3&64.5&49.9 &84.5&91.3&81.3\\
    &OwMatch~\cite{niu2024owmatch}&96.8&96.5&97.1 &71.9&80.1&63.9 &85.5&91.5&79.6\\
    \midrule
    \multirow{2}{*}{\rotatebox{90}{\tiny{\makecell{Ind.\\$\mathcal{D_L}$\\$10\%$}}}}
    &TRSSL~\cite{rizve2022towards}&92.2&94.9&89.6 &60.3&68.5&52.1 &75.4&82.6&67.8\\
    &TIDA~\cite{wang2023discover}&93.8&94.2&93.4 &65.3&73.3&56.6 &83.4&77.6&71.2\\
    \midrule
    \multirow{4}{*}{\rotatebox{90}{{\tiny{Transductive}}}}
    &ORCA~\cite{orca2022}&89.7&88.2&90.7 &48.1&66.9&43.0 &77.8&89.1&71.2\\
    &OpenNCD~\cite{ijcai2023p445}&90.1&88.4&90.6 &49.3&69.7&43.4 &81.6&90.0&77.5\\
    &OpenCON~\cite{ijcai2023p445}&90.4&89.3&91.1 &52.8&69.1&47.8 &83.8&90.6&80.8\\
    &SORL~\cite{sun2024graph}&93.5&94.0&92.5 &56.1&68.2&52.0 &-&-&-\\
    \hline
\end{tabular}
}
}
\label{tab:owssl_compare}
\end{table*}

\begin{table}[h]
\centering
\caption{Additional experiments comparison for the above selected GCD methods on CIFAR10\&100~\cite{krizhevsky2009cifar}.} 
\setlength{\tabcolsep}{3mm}{
\resizebox{0.48\textwidth}{!}{
\begin{tabular}{l|ccc|ccc}
    \hline
   \rowcolor[HTML]{D3D3D3}  &\multicolumn{3}{c|}{CIFAR10~\cite{krizhevsky2009cifar}}&\multicolumn{3}{c}{CIFAR100~\cite{krizhevsky2009cifar}}\\
\rowcolor[HTML]{D3D3D3} Method&All&Old&New&All&Old&New\\ 
    \hline \hline
    TIDA~\cite{wang2023discover}&92.9&93.3 &92.6&72.1&73.4&71.5 \\
    SORL~\cite{sun2024graph}&95.8&97.6 &95.0&76.3&79.1&70.6 \\
    OwMatch~\cite{niu2024owmatch}&\bf{96.6}&\bf{98.0} &\bf{95.9}&\bf{79.8}&\bf{81.2}&\bf{79.1} \\
    \hline
\end{tabular}
}
}
\label{tab:cifar_owssl_reimplement}
\end{table}
\section{Notes on Reproduction Settings and Protocol Differences}

As discussed in Section IV (GCD Benchmarking) of the main paper, several methods~\cite{orca2022, rizve2022openldn, rizve2022towards, ijcai2023p445, sun2024graph, wang2023discover, sun2023opencon, lps_ijcai2024, niu2024owmatch, Xiao_2024_CVPR} adopt different data-splitting protocols or backbones, leading to inconsistencies in the composition of labelled and unlabelled sets. In this section, we provide further details on these differences and explain our choice to reproduce OwMatch~\cite{niu2024owmatch} as a representative method using the commonly adopted data-splitting protocol and the DINO-pretrained ViT-Base backbone.

Several methods~\cite{orca2022, rizve2022openldn, rizve2022towards, ijcai2023p445, sun2024graph, wang2023discover, sun2023opencon, lps_ijcai2024, niu2024owmatch, Xiao_2024_CVPR} utilize ResNet-18 or ResNet-50~\cite{he2016deep} as backbone networks, in contrast to most GCD methods which adopt ViT-Base~\cite{dosovitskiy2020image} as the standard backbone. In addition, these methods follow different dataset-splitting protocols, leading to inconsistencies in the composition of labelled and unlabelled data. Several of them~\cite{rizve2022openldn, lps_ijcai2024, niu2024owmatch, rizve2022towards, wang2023discover} also adopt an inductive evaluation setting, where performance is measured on a separate, unseen test set. Among these, \cite{rizve2022towards, wang2023discover} use a 1:9 ratio for labelled-to-unlabelled data, whereas the majority of GCD methods employ a 5:5 split. In contrast, \cite{orca2022, ijcai2023p445, sun2023opencon, sun2024graph} follow a transductive evaluation protocol consistent with most GCD methods, but apply a distinct dataset-splitting strategy that results in different labelled and unlabelled sets. Notably, these four methods share the same alternative splitting protocol.

Table~\ref{tab:owssl_compare} presents a comparison of the aforementioned methods. To ensure fair benchmarking, we select the best-performing method under each splitting protocol—OwMatch~\cite{niu2024owmatch}, TIDA~\cite{wang2023discover}, and SORL~\cite{sun2024graph}—for further evaluation. Since all three originally use ResNet~\cite{he2016deep} as their backbone, we reimplement them using a ViT-Base backbone pretrained with DINO~\cite{caron2021dino} to align with the standard setup. The reimplementation results are shown in Table~\ref{tab:cifar_owssl_reimplement}. Based on these results, OwMatch~\cite{niu2024owmatch} achieves the best performance across both CIFAR-10 and CIFAR-100. Consequently, we choose OwMatch for reimplementation and evaluation on six commonly used datasets—CUB~\cite{WahCUB_200_2011}, Stanford Cars~\cite{stanfordCars}, FGVC-Aircraft~\cite{aircrafts}, CIFAR-10~\cite{krizhevsky2009cifar}, CIFAR-100~\cite{krizhevsky2009cifar}, and ImageNet-100~\cite{russakovsky2015imagenet}—as reported in the main paper.

\begin{table}[]   
    \caption{Additional illustration of detailed statistics about Dataset splitting for GCD.}
    \label{tab:datasplit_a}
    \centering
    \begin{tabular}{lc|rr|rr}
    \hline
       \rowcolor[HTML]{D3D3D3}     &    & \multicolumn{2}{c|}{Labelled}  & \multicolumn{2}{c}{Unlabelled}\\
\rowcolor[HTML]{D3D3D3}    Dataset         & Balance   & \#Image   & \#Class   & \#Image   & \#Class \\
    \hline \hline
    ImageNet-$100$~\cite{russakovsky2015imagenet}    & \cmark    & $31.9$K     & $50$        & $95.3$K     & $100$ \\
    ImageNet-$1$K~\cite{russakovsky2015imagenet}    & \cmark    & $321$K     & $500$        & $960$K     & $1000$ \\
    CIFAR-$10$~\cite{krizhevsky2009cifar}         & \cmark    & $12.5$K     & $5$         & $37.5$K     & $10$ \\
    CIFAR-$100$-$50$~\cite{krizhevsky2009cifar}        & \cmark    & $12.5$K     & $50$        & $37.5$K     & $100$ \\
    CIFAR-$100$-$20$~\cite{krizhevsky2009cifar}        & \cmark    & $20.0$K     & $80$        & $30.0$K     & $100$ \\
    CUB~\cite{WahCUB_200_2011}             & \cmark    & $1.5$K      & $100$       & $4.5$K      & $200$ \\
    Cars~\cite{stanfordCars}   & \cmark    & $2.0$K      & $98$        & $6.1$K      & $196$ \\    
    FGVC-Aircraft~\cite{aircrafts}   & \cmark    & $1.7$K      & $50$        & $5.0$K      & $100$ \\
    Oxford-Pets~\cite{oxfordPets}    & \cmark    & $0.9$K      & $19$       & $2.7$K     & $37$ \\
    Herbarium 19~\cite{Chuan19herbarium}    & \xmark    & $8.9$K      & $341$       & $25.4$K     & $683$ \\
    \hline
    \end{tabular}
\end{table}
\section{Dataset}
In this section, we present a more detailed introduction to the datasets used across different CD settings. Additional statistics on the dataset splits for GCD are provided in Table~\ref{tab:datasplit_a}.

\subsection{Generic Datasets}
\hfill \break
\textbf{ImageNet-1K~\cite{russakovsky2015imagenet} \& ImageNet-100}:
ImageNet~\cite{russakovsky2015imagenet} is a widely used dataset for natural image classification in computer vision. It contains over one million labelled images, spanning 1,000 different object categories. ImageNet-100 is constructed by randomly subsampling $100$ classes.

\hfill \break
\textbf{TinyImageNet~\cite{le2015tinyImagenet}} is a scaled-down version of the ImageNet~\cite{russakovsky2015imagenet} dataset designed for benchmarking machine learning models. It contains 200 classes, with each class represented by 500 training images and 50 validation images, totaling 100,000 images. All images are resized to 64x64 pixels, providing a challenging yet computationally efficient dataset.

\hfill \break
\textbf{CIFAR-10 \& CIFAR-100~\cite{krizhevsky2009cifar}}: CIFAR-$10$ contains $50000$ images spanning across $10$ different classes and CIFAR-$100$ includes $100$ classes, with each class containing $500$ images. All images in CIFAR-10 and CIFAR-100 are sized in $32 \times 32$.

\hfill \break
\textbf{OmniGlot~\cite{Lake15omnniglot}} comprises 1,623 handwritten characters from 50 unique alphabets, partitioned into two distinct subsets. The background set includes 964 characters from 30 alphabets, while the evaluation set consists of 659 characters from 20 alphabets. Each character represents a distinct category with 20 example images.

\hfill \break
\textbf{SVHN~\cite{Netzer2011svhn}} is a large collection of real-world digit images obtained from house number signs in Google Street View images. It includes over 70 thousand images in the training set and over 26 thousand in the test set.

\subsection{Fine-Grained Datasets}
\hfill \break
\textbf{CUB~\cite{WahCUB_200_2011}} is a widely used benchmark dataset for fine-grained visual classification tasks, particularly focused on bird species recognition. It contains 6,033 images of 200 bird species with semantic details for bird species classification.

\hfill \break
\textbf{Stanford-Cars~\cite{stanfordCars}} is a large-scale dataset designed for fine-grained vehicle classification tasks. It contains 16,185 images of 196 different car models, primarily spanning various makes, models, and years. 

\hfill \break
\textbf{FGVC-Aircraft~\cite{aircrafts}} is a fine-grained visual classification dataset focused on aircraft recognition. It contains 10,000 images spanning 100 different aircraft model variants, with each image labelled by its corresponding model. 

\hfill \break
\textbf{Oxford-Pets~\cite{oxfordPets}} is a large, fine-grained dataset designed for pet image classification and segmentation tasks. It contains 7,349 images of 37 pet breeds, including both cats and dogs, with approximately 200 images per breed.

\hfill \break
\textbf{Herbarium19~\cite{Chuan19herbarium}} is a large-scale image collection focused on plant species identification, particularly for herbarium specimen recognition. It contains over 460,000 images representing more than 32,000 species, offering a rich and diverse set of labelled plant samples.

\subsection{Domain Shift Datasets}

\hfill \break
\textbf{DomainNet~\cite{peng2019domainNet}} is a large-scale, multi-domain dataset designed for domain adaptation and transfer learning tasks. It consists of approximately 600,000 images across six distinct domains: clipart, infograph, painting, quickdraw, real, and sketch, with 345 object categories common to all domains. 

\hfill \break
\textbf{Office31~\cite{saenko2010office31}} is a widely used benchmark for domain adaptation in computer vision tasks. It consists of 4,110 images categorized into 31 classes spanning three distinct domains - Amazon, Webcam, and DSLR.

\hfill \break
\textbf{OfficeHome~\cite{venkateswara2017officehome}} is a popular benchmark for domain adaptation, consisting of 15,500 images categorized into 65 classes of everyday objects such as furniture, appliances, and office supplies. The dataset spans four different domains: Art, Clipart, Product, and Real-World.

\hfill \break
\textbf{VisDA~\cite{peng2017VisDA}} is a large-scale benchmark designed for evaluating domain adaptation algorithms, particularly in the context of image classification. It comprises over $280000$ images spanning $12$ object categories, collected from two distinct domains: synthetic renderings and real-world photographs, thereby enabling rigorous assessment of cross-domain generalization.

\subsection{Continual dataset}
\hfill \break
\textbf{iNatIGCD~\cite{zhao2023incremental}} is specifically designed for Continual Category Discovery (CCD). This dataset is based on real-world data from the iNaturalist~\cite{van2021benchmarking} platform and is structured to reflect a more natural, incremental learning environment. It features four temporal stages, each corresponding to a different geographical region and time period, providing a realistic simulation of how species data grows over time. iNatIGCD includes challenges like geographical domain shifts and fine-grained categorization, making it significantly more challenging and suitable for evaluating models under realistic conditions compared to previous artificial dataset splits. 

\subsection{Datasets for Scene Understanding}
\hfill \break
\textbf{PASCAL VOC~\cite{Pascal}}  is a benchmark in computer vision, providing annotated images across 20 object categories—such as vehicles, animals, and household items—for tasks including object detection, classification, and segmentation. 

\hfill \break
\textbf{PASCAL-$5^i$~\cite{Pascal}} is a benchmark for few-shot semantic segmentation, constructed from the PASCAL VOC 2012~\cite{Pascal} dataset.

\hfill \break
\textbf{MS COCO} is a large-scale benchmark, encompassing over $328,000$ images with more than $2.5$ million annotated instances across $91$ object categories, designed to advance object recognition and scene understanding through detailed per-instance segmentations and contextual scene information.

\hfill \break
\textbf{COCO-$20^i$~\cite{lin2015microsoftcococommonobjects}} is a few-shot semantic segmentation benchmark derived from the MS COCO~\cite{lin2015microsoftcococommonobjects} dataset, where $80$ object classes are evenly divided into four splits, each containing $20$ classes.

\subsection{Datasets for 3D}
\hfill \break
\textbf{SemanticKITTI~\cite{SemanticKITTI}} is a large-scale benchmark for semantic scene understanding in autonomous driving, providing dense point-wise annotations for over $43,000$ LiDAR scans across $22$ sequences, covering a $360$-degree field-of-view and encompassing $28$ semantic classes, including distinctions between moving and static objects.

\hfill \break
\textbf{SemanticPOSS~\cite{SemanticPOSS}} is a 3D point cloud benchmark for semantic segmentation, comprising $2988$ LiDAR scans collected in a campus environment.

\hfill \break
\textbf{ModelNet~\cite{Zhirong15CVPR}} is a large-scale benchmark for 3D object recognition, comprising over 127,000 synthetic CAD models across 662 object categories.

\hfill \break
\textbf{ShapeNetCore~\cite{chang2015shapenet}} is a richly annotated subset of the ShapeNet~\cite{chang2015shapenet} repository, encompassing approximately $51,300$ unique 3D models across $55$ common object categories, each linked to synsets in WordNet, and featuring manually verified category and alignment annotations to support research in 3D object recognition and modeling.

\hfill \break
\textbf{ScanObjectNN~\cite{uy2019revisiting}} is a real-world benchmark for 3D point cloud classification, comprising approximately 15,000 objects categorized into 15 classes with 2,902 unique object instances.

\hfill \break
\textbf{Co3D~\cite{reizenstein2021common}} is a large-scale collection designed to facilitate research in category-specific 3D reconstruction and novel view synthesis, comprising approximately $1.5$ million frames from nearly $19,000$ videos capturing objects from $50$ MS COCO categories, each annotated with camera poses and ground truth 3D point clouds.

\subsection{Datasets for Video}
\hfill \break
\textbf{Kinetics-400~\cite{kay2017kinetics}} is a large-scale benchmark for human action recognition, comprising approximately $306,245$ short video clips sourced from YouTube, each lasting around 10 seconds, categorized into $400$ distinct human action classes with at least $400$ clips per class

\hfill \break
\textbf{VGG-Sound~\cite{chen2020vggsound}} is a large-scale audio-visual collection comprising over $210,000$ videos spanning $310$ audio classes, curated from YouTube to ensure audio-visual correspondence.

\hfill \break
\textbf{UCF101~\cite{soomro2012ucf101}} is a comprehensive collection for human action recognition, comprising $13,320$ realistic video clips sourced from YouTube, categorized into $101$ distinct action classes, and encompassing a wide variety of activities such as human-object interactions, body-motion actions, human-human interactions, playing musical instruments, and sports.


\subsection{Datasets for text}
\hfill \break
\textbf{CLINC~\cite{larson-etal-2019-evaluation}} is a benchmark for intent classification in task-oriented dialogue systems, comprising $22,500$ in-domain utterances across $150$ intent classes spanning 10 domains, along with $1,200$ out-of-scope queries, designed to evaluate the performance of classifiers in identifying both supported intents and out-of-scope inputs.

\hfill \break
\textbf{HWU64~\cite{Liu2021}} is a multi-domain, fine-grained intent classification benchmark comprising $11,036$ user-generated utterances spanning 64 intents across $21$ domains related to personal assistant interactions.

\hfill \break
\textbf{WOS~\cite{8260658}} is a document classification corpus comprising $46,985$ scholarly articles categorized into $134$ subject areas under $7$ parent categories, designed to facilitate research in hierarchical text classification.

\hfill \break
\textbf{BANKING~\cite{casanueva-etal-2020-efficient}} is a fine-grained intent detection benchmark in the banking domain, comprising $13,083$ customer service queries labelled across $77$ distinct intent classes.

\hfill \break
\textbf{StackOverflow~\cite{xu-etal-2015-short}} is a comprehensive collection of user-generated content from the Stack Overflow platform, encompassing millions of programming-related questions, answers, comments, votes, tags, and badges.



\hfill \break
\textbf{TACRED~\cite{li2022openrelationeventtype}} is a large-scale benchmark for relation extraction, comprising $106,264$ sentences annotated with $41$ relation types, constructed from newswire and web text 

\hfill \break
\textbf{Re-TACRED~\cite{stoica2021re}} is a thoroughly re-annotated version of the original TACRED dataset, comprising over $91,000$ sentences across $40$ relation types, developed to address labelling inaccuracies and enhance the reliability of relation extraction model evaluations.

\hfill \break
\textbf{FewRel~\cite{han2018fewrel}} is a large-scale benchmark for few-shot relation classification, comprising $70,000$ sentences annotated with 100 distinct relations derived from Wikipedia.

%% file: tabs/CCD_dataset_statistics.tex
\begin{figure*}[h!]
    \begin{minipage}{\textwidth}
        \begin{minipage}[t]{0.72\textwidth}
        \label{tab:CCD_data}
        \centering\small
        \makeatletter\def\@captype{table}
        \caption{Statistics of the CCD benchmark datasets following the splits in Tab.~\ref{tab:CCD_data_dist}.}
        \vspace{-5pt}
        \resizebox{1.0\linewidth}{!}{
        \begin{tabular}{l|cccccccccccccc}
        \hline
          \rowcolor[HTML]{D3D3D3} 
     {Stages}
                               & \multicolumn{2}{c}{CIFAR$100$~\cite{krizhevsky2009cifar}} & \multicolumn{2}{c}{ImageNet$100$~\cite{russakovsky2015imagenet}} 
                                & \multicolumn{2}{c}{TinyImageNet~\cite{le2015tinyImagenet}} & \multicolumn{2}{c}{Caltech$101$~\cite{fei2006one}} 
                                & \multicolumn{2}{c}{Aircraft~\cite{aircrafts}} & \multicolumn{2}{c}{SCars~\cite{stanfordCars}} 
                                & \multicolumn{2}{c}{CUB~\cite{WahCUB_200_2011}}   \\ 
        
                             \rowcolor[HTML]{D3D3D3}  &\#Class  & \#Image & \#Class  & \#Image & \#Class & \#Image & \#Class  & \#Image
                                 & \#Class  & \#Image & \#Class & \#Image & \#Class & \#Image \\ \hline \hline
        
        Stage 0 $(D^{l})$        & $70$  & 30.45K & $70$  & 77.46K & $140$ & 60.90K & $71$  & 4.70K
                                & $70$  & 1.98K & $130$  & 4.62K & $140$ & 3.65K \\ 
        
        Stage 1 $(D^{u}_{1})$    & $80$  & 5.95K & $80$ & 15.14K & $160$  & 11.90K & $81$  & 0.73K
                                & $80$  & 0.37K & $152$ & 0.98K & $160$ & 0.71K \\ 
        
        Stage 2 $(D^{u}_{2})$    & $90$  & 6.55K & $90$  & 16.66K & $180$ & 13.10K & $91$ & 0.65K
                                & $90$ & 0.43K & $174$  & 1.13K & $180$  & 0.79K \\ 
        
        Stage 3 $(D^{u}_{3})$    & $100$ & 7.05K & $100$ & 17.94K & $200$ & 14.10K & $101$  & 1.12K
                                & $100$  & 0.55K & $196$ & 1.38K & $200$  & 0.85K \\ \bottomrule
        \end{tabular}
        }
        \end{minipage}
        \begin{minipage}[t]{0.3\textwidth}
        \centering\small
        \makeatletter\def\@captype{table}
        \caption{Data splits for CCD.}
        \label{tab:CCD_data_dist}
        \vspace{-5pt}
        \resizebox{0.98\linewidth}{!}{
        \begin{tabular}{l|cccc} 
            \hline
          \rowcolor[HTML]{D3D3D3}   \multicolumn{1}{c|}{\large Class splits} & \large $D^{l}$ & \large $D^{u}_{1}$ & \large $D^{u}_{2}$ & \large $D^{u}_{3}$ \\ [0.5ex] 
            \hline\hline
            $\{y_i \mid y_i \leq 0.7 \ast |\mathcal{Y}| \}$ 
            & $87\%$  & $7\%$  & $3\%$  & $3\%$ \\ [0.5ex] \hline
            $\{y_i \mid 0.7 \ast |\mathcal{Y}| < y_i \leq 0.8 \ast |\mathcal{Y}| \}$ 
            & 0\% & $70\%$  & $20\%$  & $10\%$  \\ [0.5ex] \hline
            $\{y_i \mid 0.8 \ast |\mathcal{Y}| < y_i \leq 0.9 \ast |\mathcal{Y}| \}$ 
            & 0\% & 0\% & $90\%$  & $10\%$  \\ [0.5ex] \hline
            $\{y_i \mid 0.9 \ast |\mathcal{Y}| < y_i \leq |\mathcal{Y}| \}$ 
            & 0\% & 0\% & 0\% & $100\%$  \\ [0.5ex] \bottomrule
        \end{tabular}
        }
        \end{minipage}
    \end{minipage}
\end{figure*}

%% file: tabs/benchmark_ccd.tex
\begin{table*}[ht]
\centering
\caption{Comparison of state-of-the-art CCD methods. Results are reported in $cACC$ across `All', `Old' and `New' categories. }
\label{tab:ccd_results}
\setlength{\tabcolsep}{1.3mm}{
\resizebox{\textwidth}{!}{
\begin{tabular}{l||ccc|ccc|ccc|ccc|ccc|ccc|ccc}
\hline
\rowcolor[HTML]{D3D3D3}& \multicolumn{3}{c|}{CIFAR-100} & \multicolumn{3}{c|}{ImageNet-100} & \multicolumn{3}{c|}{TinyImageNet} & \multicolumn{3}{c|}{Caltech-101} & \multicolumn{3}{c|}{FGVC-Aircraft} & \multicolumn{3}{c|}{Stanford-Cars} & \multicolumn{3}{c}{CUB} \\
\rowcolor[HTML]{D3D3D3}Method&  {All} &  {Old} &  {New} &  {All} &  {Old} &  {New} &  {All} &  {Old} &  {New} &  {All} &  {Old} &  {New} &  {All} &  {Old} &  {New} &  {All} &  {Old} &  {New} &  {All} &  {Old} &  {New} \\ 
\hline \hline
ORCA$^\dagger$~\cite{orca2022} & \underline{60.91} & 66.61 & \underline{58.33} & 40.29 & 45.85 & 35.40 & {54.71} & 63.13 & \underline{51.93} & 76.77 & 82.80 & 73.20 & 30.77 & 25.71 & 32.44 & 20.79 & 33.40 & 17.60 & 41.73 & 66.19 & 34.14 \\
GCD$^\dagger$~\cite{vaze2022gcd} & 58.18 & 72.27 & 52.83 & \underline{69.41} & 81.56 & 65.65 & 55.20 & 65.87 & 51.61 & 78.27 & 86.60 & 72.92 & 47.37 & \underline{61.43} & \underline{42.53} & 39.21 & 58.29 & 33.45 & \underline{54.98} & 75.47 & \underline{48.15} \\
SimGCD$^\dagger$~\cite{wen2023simgcd}& 25.56 & 38.76 & 20.43 & 31.38 & 40.47 & 27.44 & 33.40 & 29.11 & 34.74 & 33.65 & 37.53 & 31.62 & 29.03 & 35.72 & 25.61 & 21.01 & 40.93 & 16.48 & 39.89 & 59.25 & 33.75 \\
GM$^\dagger$~\cite{zhang2022growmergeunifiedframework} & 57.43 & 63.68 & 55.31 & 67.84 & 75.10 & \underline{66.60} & 52.14 & 59.68 & 49.96 & 75.75 & 83.66 & 71.59 & 31.06 & 33.33 & 30.78 & 21.90 & 35.29 & 18.17 & 38.87 & 65.00 & 30.29 \\
MetaGCD$^\dagger$~\cite{wu2023metagcd} & 55.49 & 69.38 & 48.98 & 66.41 & 80.54 & 60.65 & \underline{55.26} & 66.12 & 50.79 & \underline{80.75} & 89.02 & \underline{75.86} & 44.63 & 59.05 & 39.39 & 35.98 & 56.97 & 29.96 & 44.59 & 74.40 & 35.40 \\
PA-CGCD$^\dagger$~\cite{Kim_2023_pacgcd}& 58.25 & \bf{87.11} & 49.04 & 64.79 & \bf{91.15} & 57.83 & 51.13 & \bf{74.95} & 43.52 & 77.96 & \bf{94.75} & 69.66 & \underline{48.24} & \bf{73.09} & 40.60 & \bf{48.88} & \bf{80.43} & \underline{33.54} & 52.48 & \bf{77.26} & 44.74 \\
PromptCCD~\cite{cendra2024promptccd} &  \bf{64.17} & \underline{75.57} & \bf{60.34} &  \bf{76.16} & \underline{81.76} & \bf{74.35} &  \bf{61.84} & \underline{66.54} & \bf{60.26} &  \bf{82.44} & \underline{89.08} & \bf{79.72} &  \bf{52.64} & 60.48 & \bf{50.23} &  \underline{44.07} & \underline{66.36} & \bf{36.83} &  \bf{55.45} & \underline{75.48} & \bf{48.56} \\
\hline
\end{tabular}%
}
}
\RaggedRight
\noindent\footnotesize{$^\dagger$ denotes results from \cite{cendra2024promptccd}.}
\vspace{-15pt}
\end{table*}